\PassOptionsToPackage{unicode}{hyperref}
\PassOptionsToPackage{hyphens}{url}
\documentclass[
]{article}
\usepackage{amsmath,amssymb}
\usepackage{lmodern}
\usepackage{float}
\usepackage[T1]{fontenc}
\usepackage[utf8]{inputenc}
\usepackage{textcomp}
\usepackage{upgreek}
\usepackage{bm}

\IfFileExists{upquote.sty}{\usepackage{upquote}}{}
\IfFileExists{microtype.sty}{% use microtype if available
  \usepackage[]{microtype}
  \UseMicrotypeSet[protrusion]{basicmath} % disable protrusion for tt fonts
}{}
\makeatletter
\@ifundefined{KOMAClassName}{% if non-KOMA class
  \IfFileExists{parskip.sty}{%
    \usepackage{parskip}
  }{% else
    \setlength{\parindent}{0pt}
    \setlength{\parskip}{6pt plus 2pt minus 1pt}}
}{% if KOMA class
  \KOMAoptions{parskip=half}}
\makeatother
\usepackage{xcolor}
\usepackage{longtable,booktabs,array}
\usepackage{multirow}

\usepackage[a4paper,margin=1in]{geometry}   % wider, symmetric text area (use letterpaper if you prefer)
\usepackage{float}                           % [H] = keep float exactly here
\usepackage{tabularx}
\usepackage[font=small,labelfont=bf,labelsep=period]{caption}
\newcolumntype{C}[1]{>{\centering\arraybackslash}m{#1}}

\usepackage{calc} % for calculating minipage widths
\usepackage{etoolbox}
\makeatletter
\patchcmd\longtable{\par}{\if@noskipsec\mbox{}\fi\par}{}{}
\makeatother
\IfFileExists{footnotehyper.sty}{\usepackage{footnotehyper}}{\usepackage{footnote}}
\makesavenoteenv{longtable}
\usepackage{graphicx}
\makeatletter
\def\maxwidth{\ifdim\Gin@nat@width>\linewidth\linewidth\else\Gin@nat@width\fi}
\def\maxheight{\ifdim\Gin@nat@height>\textheight\textheight\else\Gin@nat@height\fi}
\makeatother
\setkeys{Gin}{width=\maxwidth,height=\maxheight,keepaspectratio}
\makeatletter
\def\fps@figure{htbp}
\makeatother
\ifLuaTeX
  \usepackage{selnolig}  % disable illegal ligatures
\fi
\IfFileExists{bookmark.sty}{\usepackage{bookmark}}{\usepackage{hyperref}}
\IfFileExists{xurl.sty}{\usepackage{xurl}}{} % add URL line breaks if available
\hypersetup{
  hidelinks,
  pdfcreator={LaTeX via pandoc}}

\author{}
\date{}
\begin{document}

\begin{center}
  {\Large\bfseries Learning to Orchestrate Evolutionary Search:\\ Progression-Aware Deep Reinforcement Learning for Dynamic DE-CMA-ES Coordination in Optimization and Structural Model Updating\par}
  \vspace{1em}
  {\small Lechen Li\textsuperscript{a,b,}\footnote{Corresponding author,
    E-mail address: LechenLi@um.edu.mo}, \;
   Rongye Shi\textsuperscript{c}, \;
   Wanhuan Zhou\textsuperscript{a}\par}
  \vspace{0.6em}
  {\footnotesize
    \emph{\textsuperscript{a}State Key Laboratory of Internet of Things for Smart City, University of Macau, Macau 999078, China}\par
   \emph{\textsuperscript{b}College of Water Conservancy and Hydropower Engineering, Hohai University, Nanjing 210098, China}\par
   \emph{\textsuperscript{c}School of Artificial Intelligence, Beihang University, Beijing 100191, China}\par}
\end{center}
\vspace{0.5em}

\textbf{Abstract:} Solving high-dimensional structural model updating
problems requires an algorithm capable of navigating complex, non-convex
landscapes with correlated parameters. Existing hybrid evolutionary
algorithms typically rely on static architectures or fixed switching
rules, resulting in disjointed search phases. To address this, this
study proposes a Deep Reinforcement Learning-governed dynamic DE-CMA-ES
Orchestration (DRL-DCO) algorithm, in which a Deep Deterministic Policy
Gradient (DDPG)-based actor-critic agent continuously governs the
evolutionary process as a single, unified system rather than a
mechanical concatenation of algorithms. Guided by a progression-aware
state representation and a diversity-informed reward, the agent fluidly
reallocates computational resources between the difference-vector-based
exploration of Differential Evolution (DE) and the covariance-guided
exploitation of CMA-ES, while jointly regulating population size, elite
preservation, and a restart mechanism to escape local optima. This
allows DRL-DCO to autonomously transition between exploration-dominant,
exploitation-dominant, and mixed-strategy regimes across generations.
Beyond the training phase, the trained actor can operate in a supervision-free inference mode, where the internalized policy autonomously orchestrates DE and CMA-ES control from observed search states through forward inference alone, without critic evaluation or weight updates, enabling faster deployment while retaining full effectiveness. Validated on high-dimensional single-objective optimization benchmarks and the IASC-ASCE structural health monitoring benchmark, DRL-DCO achieves superior convergence accuracy and robustness compared to state-of-the-art adaptive and hybrid evolutionary algorithms, as well as single-operator DRL-governed
baselines.


\hypertarget{introduction}{%
\section{1. Introduction}\label{introduction}}

\hypertarget{background-and-existing-studies}{%
\subsection{1.1 Background and existing
studies}\label{background-and-existing-studies}}

Structural model updating, as an essential back-analysis strategy within
Structural Health Monitoring (SHM), has been studied for decades for its
effectiveness in assessing a structural system\textquotesingle s
condition through tracking key physical parameters. In recent years, the
field has continued to evolve in response to the growing population of
aging structures requiring condition assessment, and, more importantly,
in response to continually and explosively advancing data-driven and
Artificial Intelligence (AI)-driven techniques {[}1, 2{]}.

One of the earliest and most influential stages in the evolution of
structural model updating methods centers on optimization, since most
model updating and related back-analysis tasks are fundamentally
minimization problems over the discrepancy between a model's predicted
response and the observed response of the target structure {[}3{]}. This
framework is typically built upon a physical model, e.g., a Finite
Element Model (FEM), whose structural parameters are optimized until the
simulated output matches the observations, giving rise to the well-known
Finite Element Model Updating (FEMU) framework {[}4, 5{]}. Accordingly,
a wide range of optimization algorithms has been extensively studied.
Evolutionary Algorithms (EAs) were among the earliest and most
thoroughly investigated optimizer classes {[}6{]}, with representative
examples including Genetic Algorithms (GA) {[}7{]}, Particle Swarm
Optimization (PSO) {[}8{]}, Differential Evolution (DE) {[}9{]},
Covariance Matrix Adaptation Evolution Strategy (CMA-ES) {[}10{]}, and
their many subsequent variants and hybrids {[}11{]}. EAs offer engineers
straightforward implementations and intuitive search logic within the
FEMU framework, but they also present practical challenges: model
evaluations can be computationally expensive, and hyperparameter
calibration typically demands considerable empirical experience, since
properly balancing exploration and exploitation to locate the global
optimum while avoiding local minima remains inherently difficult
{[}12{]}.

Beyond EA-based model updating, which optimizes structural parameters
directly and deterministically, Bayesian model updating offers a
probabilistic and more information-rich alternative that fully
incorporates observation data: a prior distribution over the unknown
model parameters (informed by expert judgment) is updated to a posterior
distribution once the observations are taken into account {[}13{]}.
Classical approaches to estimating these posteriors are dominated by
Monte Carlo methods, including Markov Chain Monte Carlo (MCMC) {[}14{]},
Transitional MCMC (TMCMC) {[}15{]}, and particle filter-based approaches
(also known as Sequential Monte Carlo) {[}16{]}. Recent studies cover
some relatively more advanced approaches such as those based on
variational Bayesian inference {[}17, 18{]}, which are generally based
on approximating the target posterior density with a variational density
searched from a rich family of parameterized densities by minimizing the
Kullback-Leibler (KL) divergence between the true posterior density and
the variational one.

With the rapid growth of AI, AI-driven methods have become increasingly
prominent across engineering fields, and data-driven surrogate modeling
has emerged as another major thrust in structural model updating
research {[}19{]}. Surrogate models have strong potential to reduce
reliance on high-level FEM expertise and to lower the computational cost
of repeated FEM evaluations, thereby facilitating practical FEMU
implementation {[}20{]}. Neural network-based surrogates in particular,
enable efficient inverse analysis and rapid, adaptive parameter
identification {[}20{]}. Sabamehr et al. {[}21{]} combined neural
networks with GAs to relate natural frequency shifts to sectional
property changes in bridges, outperforming traditional matrix-based FEMU
methods. Yin et al. {[}22{]} developed a Bayesian neural network for
FEMU of a pedestrian bridge that achieved greater accuracy and
efficiency than finite-difference approaches. Zhang et al. {[}23{]}
introduced a physics-guided neural network integrating data-driven
learning with FEM principles to improve damage detection in both
simulated and experimental settings. Despite these advances, existing
data-driven surrogates still fall short of fully integrating physical
principles with measured response data, and cannot yet reliably
substitute for FEMU or other physics-based models in capturing the full
complexity of structural dynamic behavior {[}17{]}. This gap has also
been observed in many advanced Physics-Informed Neural Networks (PINNs),
which, despite explicitly embedding physical constraints into the
learning process, continue to struggle under ambiguous boundary
conditions or noisy measurements {[}24{]}. These limitations, however,
also mark out a broad and promising direction for further development of
data-driven surrogate modeling.

This AI-driven mindset in model updating has so far concentrated mainly
on data-driven surrogates, with comparatively less attention paid to the
optimization side of the problem. One representative optimization
strategy that does incorporate AI is Bayesian Optimization (BO), which
uses an acquisition function to decide where to sample next, generally
in conjunction with a data-driven surrogate such as Gaussian Process
models {[}25, 26{]}. Notably, many of these BO methods accelerate
convergence and reduce the number of required FEM evaluations, but
remain fundamentally a data-driven (probabilistic) surrogate-based model
updating framework rather than a direct or efficient advance in
optimization strategies.

More recently, Reinforcement Learning (RL), emerging from optimal
control theory and increasingly driven by AI, has been extensively
developed and applied across engineering fields, including SHM {[}27{]}.
Among them, Deep Reinforcement Learning (DRL) {[}28{]}, as the
state-of-the-art integration of RL with deep learning, has become a
leading paradigm for agent-based learning and has been actively explored
for developing intelligent, manual-operation-free optimization
algorithms, showing clear advantages over traditional optimization
methods in dynamic and complex environments. Within SHM, Meng et al.
{[}29{]} developed a Double Deep Q-Network (DDQN)-based DRL algorithm to
solve the discrete combinatorial optimization problem of vibration
sensor placement. Zhong et al. {[}30{]} proposed a heuristic-enhanced
DRL algorithm for efficient optimal placement of multiple sensor types.
Guo et al. {[}31{]} proposed a DRL-based control strategy for the
dynamic response of building structures during earthquakes and other
natural disasters. Nishat et al. {[}32{]} proposed a Proximal Policy
Optimization (PPO)-based DRL framework for autonomous control of
Wireless Sensor Networks (WSNs) tailored to SHM applications. However,
existing studies remain limited in developing DRL-based optimization
algorithms suited to direct mathematical optimization and structural
model updating tasks, which are inherently continuous-variable problems.

For directly solving mathematical optimization problems, a growing body
of work has explored integrating EA and RL mechanisms into a new
direction termed Evolutionary Reinforcement Learning (ERL) {[}33{]},
which has shown notable performance gains. Two primary sub-directions
stand out: EA-assisted optimization of RL {[}33{]} , where the EA's
population-based evolutionary mechanism generates a more diverse set of
experiences to improve DRL training, and RL-assisted optimization of EAs
{[}34{]} , where RL dynamically guides the EA\textquotesingle s search
and refinement process. The RL-assisted EA direction still faces notable
challenges. First, generalization remains largely unresolved, since many
methods are developed for specific target problems, constrained by EA
operators suited only to a single problem type. Second, many existing
methods offer only limited control over key hyperparameters, typically
restricted to mutation and crossover rates {[}35{]}, because designing
RL-assisted dynamic hyperparameter outputs from the actor model requires
a deep understanding of both the EA's internal mechanisms and the target
problem's properties {[}34{]}. Moreover, many ERL or DRL-based methods
remain focused on purely mathematical or control benchmarks with limited
real-world engineering application, and are further hindered by heavy
reliance on supervised-style feedback, large EA populations, and long
per-episode training. These limitations can leave them without
efficient, generalizable inference once training ends, often making them
poorly suited to real-world model updating tasks that already carry the
added cost of neural network training and/or physical simulation on top
of the optimization itself.

\hypertarget{research-focus-and-key-contributions-of-this-study}{%
\subsection{1.2 Research focus and key contributions of this
study}\label{research-focus-and-key-contributions-of-this-study}}

This study is situated within the domain of RL-assisted optimization of EAs and introduces an innovative DRL-governed dynamic DE–CMA-ES Orchestration (DRL-DCO) algorithm. The DRL-DCO algorithm employs an actor-critic orchestration agent, built upon the Deep Deterministic Policy Gradient (DDPG) method [36], to dynamically coordinate the DE and CMA-ES search mechanisms through a multi-dimensional continuous action space governing key hyperparameters and orchestration decisions, guided by a progression-aware state representation and a diversity-informed reward function. Whereas many existing RL-assisted EA approaches confine the RL component to intermittent intervention on a single evolutionary strategy and/or limited landscape or progression awareness, DRL-DCO's actor-critic agent maintains continuous, progression-aware oversight of two parallel evolutionary processes throughout the search. This persistent, multi-dimensional governance represents a more thorough realization of the RL-assisted EA paradigm. The DRL-DCO algorithm is evaluated across high-dimensional single-objective optimization benchmarks and a practical FEMU task, spanning both synthetic optimization landscapes and practical structural engineering generalization. The key contributions are summarized as follows.

1) Rather than hard-switching between algorithms, the DRL-DCO algorithm
performs continuous orchestration of DE and CMA-ES through a
compositional control mechanism governed by a continuously learnable
balancing parameter. This parameter allocates search effort between the
two operators on the non-elite subset of each generation, preserving the
global distributional structure tracked by CMA-ES through its covariance
model while retaining the aggressive, direction-agnostic local search of
DE. By adaptively regulating this balance, the agent reduces
over-aggressive search steps and mitigates premature convergence.

2) The DRL-DCO algorithm is equipped with an enhanced landscape-informed
state representation that augments the candidate solution vector with
additional progression-aware descriptors, including population
diversity, distributional shape statistics, and convergence-rate
indicators. This enriched state gives the actor a comprehensive,
temporally coherent view of the search dynamics and geometry, enabling
more informed and timely adjustment of DE and CMA-ES hyperparameters
across generations.

3) To further enhance adaptability, the DRL-DCO algorithm incorporates a
learnable population-scaling factor together with an elite-preservation
mechanism, jointly adjusting the effective population size and the
proportion of elite solutions carried forward across generations based
on the current search state. This allows the agent to deploy smaller,
computationally efficient populations during exploitation-dominant
phases and expand population coverage during exploration-demanding
phases, improving both efficiency and stability.

4) The DRL-DCO algorithm incorporates a learned restart policy governed
by a learnable restart probability, estimated from stall-count and
diversity statistics within the state representation, which determines
when the population should be reinitialized to escape deceptive local
minima on highly multimodal landscapes. This restart policy is
complemented by two auxiliary, independently learnable CMA-ES control
actions, a step-size parameter and a covariance-scaling action, which
continuously regulate the search radius and distribution shape across
generations to support the agent\textquotesingle s broader
exploration--exploitation balance.

5) The DRL-DCO algorithm operates through two distinct phases: an
agent-training phase, in which the actor--critic agent learns its
orchestration policy through online interaction under critic-mediated
reward supervision, and an agent-inference phase, in which the trained
actor operates autonomously via forward inference alone, requiring no
critic evaluation, network updates, or reward feedback. This design enables the learned search strategy to be reused across different optimization runs without online policy adaptation, providing a practical route toward efficient and automated DRL-assisted evolutionary optimization.

\hypertarget{methodology}{%
\section{2. Methodology}\label{methodology}}

The DRL-DCO algorithm reformulates evolutionary optimization as a
controlled stochastic process, in which the DE and CMA-ES operators are
adaptively regulated by a learned control policy rather than a static
architecture or predefined switching rule. Here, the DRL-DCO algorithm
refers to the overall optimization framework developed for both FEMU and
general single-objective optimization problems, integrating dynamic
DE--CMA-ES orchestration, population evolution, and RL-based adaptive
search. Embedded within this framework, the DRL agent (specifically, the
DDPG-based actor--critic agent introduced in Section 1.2) serves as the
decision-making component, learning a control policy that regulates the
evolutionary process through state observation, action generation, and
reward feedback.

From this perspective, the optimization process is a sequence of
population transformations driven by a learnable control policy, of
which classical EAs represent a special case with a static or
heuristic-based policy. In the DRL-DCO algorithm, this policy instead
observes population-level statistics, fitness-progression signals, and
convergence indicators to jointly orchestrate DE's
difference-vector-driven search and CMA-ES\textquotesingle s
covariance-based search within a single, unified process.

The remainder of this section develops this formulation of the DRL-DCO
algorithm in detail: Section 2.1 first presents the mathematical
formulation of the FEMU problem, followed by the controlled evolutionary
process view (Section 2.2), the DRL core mechanism (Section 2.3), the
resulting dynamic DE--CMA-ES orchestration mechanism (Section 2.4), and
the algorithm implementation details (Section 2.5).

\hypertarget{problem-formulation-of-femu}{%
\subsection{2.1 Problem formulation of
FEMU}\label{problem-formulation-of-femu}}

The FEMU is commonly formulated as a parameter identification problem in
which uncertain model parameters are inferred by minimizing the
discrepancy between simulated structural characteristics and measured
data. Let \(\boldsymbol{\theta} \in \mathbb{R}^{N_{\text{d}}}\) denote the vector
of updating parameters, where \(N_{\text{d}}\) is the problem dimensionality;
these may represent stiffness modifiers, material properties, mass
scaling factors, or other physically meaningful quantities within a
finite element or reduced-order model. In this study, the updating
parameters are chosen as stiffness-related coefficients since stiffness
degradation is generally the dominant contributor to structural damage.
A composite objective function is adopted that combines the Mean Squared
Error (MSE) of natural frequencies with a Modal Assurance Criterion
(MAC)-based error of mode shapes, a formulation widely adopted in the
FEMU literature {[}4, 5{]}:

\[\operatorname{f}\left( \boldsymbol{\theta} \right) = w_{\text{f}} \cdot \frac{1}{N_{\text{f}}}\sum_{i = 1}^{N_{\text{f}}}\left( \frac{\omega_{i}^{\text{sim}}\left( \boldsymbol{\theta} \right) - \omega_{i}^{\exp}}{\omega_{i}^{\exp}} \right)^{2} + w_{\text{m}} \cdot \frac{1}{N_{\text{m}}}\sum_{j = 1}^{N_{\text{m}}}\left( 1 - MAC\left( \phi_{j}^{\text{sim}}\left( \boldsymbol{\theta} \right),\phi_{j}^{\exp}\  \right) \right)\]

where \(\boldsymbol{\theta}\) is the vector of model parameters to be
updated; \(\omega_{i}^{\text{sim}}\left( \boldsymbol{\theta} \right)\) and
\(\omega_{i}^{\exp}\)\hspace{0pt} are the simulated and measured \(i\)th
natural frequencies; \(\phi_{j}^{\text{sim}}\left( \boldsymbol{\theta} \right)\)
and \(\phi_{j}^{\exp}\) are the simulated and measured \(j\)th mode
shapes, with \(N_{\text{f}}\) and \(N_{\text{m}}\)\hspace{0pt} being the numbers of
considered frequencies and mode shapes, respectively;
\(w_{\text{f}}\)\hspace{0pt} and \(w_{\text{m}}\)\hspace{0pt} are weighting
coefficients for frequency and mode shape errors, each set to 0.5 to
ensure balanced contributions. The goal is to determine
\(\boldsymbol{\theta}^{*}\) that minimizes the objective function
\(\operatorname{f}\left( \boldsymbol{\theta} \right)\).

The resulting optimization problem is high dimensional and exhibits
strong parameter coupling induced by structural dynamics, such that
multiple, physically distinct parameter combinations may yield similar
modal responses. This non-uniqueness, combined with modeling
uncertainty, gives rise to a nonconvex, multimodal objective landscape.
Consequently, even where gradient information can be derived,
gradient-based methods can remain prone to convergence toward local
optima and is highly sensitive to the initial parameter guess. These
motivate the use of gradient-free, population-based optimization
strategies capable of exploring the parameter space broadly and locating
the global optimum without reliance on gradient or prior information of
parameters.

This study addresses the FEMU problem within a two-stage,
modal-property-based updating framework: rather than directly matching
time or frequency domain structural responses, the updating process
minimizes discrepancies in modal properties obtained through eigenvalue
analysis, substantially reducing the computational burden of repeated
finite element evaluations while retaining the dynamic information
essential for reliable parameter identification. The present study
focuses on this second-stage optimization problem and the proposed
DRL-DCO solver, without addressing the preceding system identification
procedure.

\hypertarget{evolutionary-optimization-as-a-controlled-stochastic-process}{%
\subsection{2.2 Evolutionary Optimization as a Controlled Stochastic
Process}\label{evolutionary-optimization-as-a-controlled-stochastic-process}}

Let
\(\boldsymbol{\chi}^{(t)} = \{\boldsymbol{x}_{1}^{(t)},\boldsymbol{x}_{2}^{(t)},\ \ldots,\ \boldsymbol{x}_{N_{\text{s}}}^{(t)}\}\)
denote a population of \(N_{\text{s}}\) candidate solutions at generation
\(t\), where each solution 
\(\boldsymbol{x}_{k}^{(t)} \in \mathbb{R}^{N_{\text{d}}}\) is an
\(N_{\text{d}}\)-dimensional vector (\(k = 1,\ldots,N_{\text{s}}\)). In a general
evolutionary optimization framework, the population evolves according
to a transformation:

\[\boldsymbol{\chi}^{(t+1)} = \operatorname{T}_{\text{e}}\left(\boldsymbol{\chi}^{(t)};\,\boldsymbol{\theta}\right)\]

where the operator \(\operatorname{T}_{\text{e}}( \cdot )\) represents the evolutionary
operations and effects, including mutation, recombination, selection,
and adaptation mechanisms, and \(\boldsymbol{\theta}\) denotes the
hyperparameters governing them.

In conventional EAs, \(\operatorname{T}_{\text{e}}( \cdot )\) and \(\boldsymbol{\theta}\) are
fully specified before the optimization begins. That is, all
hyperparameters governing mutation strength, recombination rates, step
sizes, and population size remain fixed or are adjusted through
predefined heuristic rules. As a result, the search dynamics are largely
decoupled from the actual landscape encountered during optimization.

In this study, the DRL-DCO algorithm instead reformulates this
evolutionary process as a controlled stochastic system:

\[\boldsymbol{\chi}^{(t+1)} = \operatorname{T}_{\text{s}}\left(\boldsymbol{\chi}^{(t)}\middle|\boldsymbol{a}^{(t)}\right),
\qquad \boldsymbol{a}^{(t)} = \pi\left(\boldsymbol{s}^{(t)}\right)\]

where \(\boldsymbol{a}^{(t)}\) denotes an action (control) vector applied at
generation \(t\), and \(\boldsymbol{s}^{(t)}\) is the state observed by the
agent, constructed from \(\boldsymbol{\chi}^{(t)}\). The action vector
explicitly parameterizes the evolution operator, allowing its behavior
to change dynamically as optimization progresses. Under this
formulation, the operator \(\operatorname{T}_{\text{s}}( \cdot )\) plays the role of the
stochastic system dynamics; the population \(\boldsymbol{\chi}^{(t)}\)
represents the underlying system state; and the action vector
\(\boldsymbol{a}^{(t)}\) regulates exploration intensity, exploitation
strength, and structural adaptation of the search. This controlled
process view provides a natural bridge between evolutionary
optimization and optimal control theory, and forms the basis for the DRL
core mechanism introduced in Section 2.3, which learns the control
policy \(\pi\) generating \(\boldsymbol{a}^{(t)}\).

\hypertarget{markov-decision-process-formulation-for-the-optimization}{%
\subsection{2.3 Markov Decision Process Formulation for the
Optimization}\label{markov-decision-process-formulation-for-the-optimization}}

To enable adaptive, state-dependent control of the evolutionary search
process, the optimization process introduced in Section 2.2 is
formalized as a Markov Decision Process (MDP) {[}37{]}. This formulation
provides a mathematically principled mechanism for learning an optimal
control policy that governs the evolution operator based on the observed
search state. The MDP is defined by a state space, an action space,
transition dynamics (inherited from the population evolution operator
\(\operatorname{T}_{s}( \cdot )\) in Section 2.2), and a reward function, each
specified for the DRL-DCO algorithm in the following subsections. The
optimization proceeds over successive generations \(t\), with the FEMU
(or general optimization objective function)
\(\operatorname{f}\left( \boldsymbol{\theta} \right)\) providing the basis for evaluating
population fitness and constructing the reward signal at each
generation.

\hypertarget{state-space}{%
\subsubsection{2.3.1 State Space}\label{state-space}}

The state at generation \(t\), denoted \(\boldsymbol{s}^{(t)}\), is
constructed to characterize both the current search quality and the
evolutionary dynamics of the population. Rather than relying solely on
objective function values, the state representation integrates
normalized performance indicators and population-level geometric
descriptors that reflect the search status in a scale-consistent manner.
The state vector is defined as:

\[\boldsymbol{s}^{\mathbf{(}t\mathbf{)}} = \left\lbrack f_{\text{best}}^{(t)},f_{\text{mean}}^{(t)},f_{\text{std}}^{(t)},f_{\text{div}}^{(t)},f_{\text{ng}}^{(t)},f_{\text{imp}}^{(t)},\sigma_{1}^{(t)},\ldots,\sigma_{d}^{(t)},...,\sigma_{N_{\text{d}}}^{(t)} \right\rbrack\]

where \(f_{\text{best}}^{(t)}\) denotes the normalized best fitness achieved so
far, representing the global search performance; \(f_{\text{mean}}^{(t)}\) and
\(f_{\text{std}}^{(t)}\) are the normalized mean and standard deviation of the
population fitness, describing overall population quality and
dispersion; \(f_{\text{div}}^{(t)}\) measures population diversity, defined as
the average distance of individuals to the population centroid;
\(f_{\text{ng}}^{(t)}\) is a normalized generation index, scaled by the maximum
number of generations per episode, providing temporal information on
search progress; \(f_{\text{imp}}^{(t)}\) captures recent fitness improvement,
defined as
\(f_{\text{imp}}^{(t)} = \max\{ 0,\ \ f_{\text{best}}^{(t-1)} - f_{\text{best}}^{(t)}\}\),
serving as a convergence indicator; \(\sigma_{d}^{(t)}\) is the
normalized standard deviation of the population along each parameter
dimension (\(d=1,\ldots,N_{\text{d}}\)), capturing anisotropic spread of the
search.

This state construction provides an informative description of the
optimization landscape, capturing both convergence status and population
geometry. In particular, the diversity and improvement indicators allow
the agent to adaptively balance exploration and exploitation through its
action outputs, while the dimension-wise spread supports directionally
sensitive (anisotropic) search regulation.

\hypertarget{action-vector-and-control-variables}{%
\subsubsection{2.3.2 Action Vector and Control
Variables}\label{action-vector-and-control-variables}}

The action \(\boldsymbol{a}^{(t)}\) at generation \(t\) is a continuous
control vector that regulates the evolutionary operator, defined as:

\[\boldsymbol{a}^{(t)} = \lbrack F^{(t)},{CR}^{(t)},\lambda^{(t)},{LR}_{\text{cov}}^{(t)},\Delta\sigma^{(t)},p_{\text{res}}^{(t)},r_{\text{eli}}^{(t)},\beta^{(t)}\rbrack\]

where each component governs a specific aspect of the evolutionary
search: \(F^{(t)}\)and \({CR}^{(t)}\) control the mutation scaling
factor and crossover probability of the DE operator; \(\lambda^{(t)}\)
scales the effective population size relative to a base population;
\({LR}_{\text{cov}}^{(t)}\) is the learning rate of the CMA-ES
covariance matrix update, governing how rapidly the search distribution
adapts; \(\Delta\sigma^{(t)}\)introduces a bias term to the CMA-ES
step-size adaptation mechanism, directly regulating the global search
radius; \(p_{\text{res}}^{(t)}\) specifies the restart probability applied to
newly generated individuals, supporting escape from local minima;
\(r_{\text{eli}}^{(t)}\) determines the proportion of elite individuals
preserved across generations, strengthening exploitation; and
\(\beta^{(t)} \in \lbrack 0,1\rbrack\) governs the allocation ratio
between DE-based and CMA-ES-based offspring generation.

This action formulation enables continuous, coordinated orchestration of
multiple evolutionary mechanisms within each generation, rather than
discrete switching between algorithms, allowing the search dynamics to
transition smoothly among exploration-dominant, exploitation-dominant,
and hybrid regimes as optimization progresses across the FEMU or
general-optimization landscape.

\hypertarget{reward-function}{%
\subsubsection{2.3.3 Reward Function}\label{reward-function}}

The reward \(r^{(t)}\) is designed to quantify how effectively the
applied action \(\boldsymbol{a}^{(t-1)}\) advances the optimization process at generation \(t\),
combining improvement-driven reinforcement, diversity encouragement, and
stagnation penalization:

\[r^{(t)} = \ \lambda_{1} \cdot \max(0,\ f_{\text{best}}^{(t-1)}\  - \ f_{\text{best}}^{(t)}) + \lambda_{2} \cdot P_{\text{std}}^{(t)} - \ \lambda_{3} \cdot C_{\text{sta}}^{(t)} + \ \mathbb{I}(f_{\text{best}}^{(t)}\  < \ \varepsilon)\]

where \(f_{\text{best}}^{(t)}\) denotes the best fitness value in the
current population; \(P_{\text{std}}^{(t)}\) measures population diversity, computed as the per-dimension standard deviation of the population, averaged across all \(N_{\text{d}}\) dimensions at generation \(t\); \(C_{\text{sta}}^{(t)}\) is a stall counter recording consecutive generations without improvement; and \(\mathbb{I}( \cdot )\) is an indicator function providing a terminal reward once the solution accuracy reaches a predefined threshold \(\varepsilon\). The non-negative coefficients \(\lambda_{1},\ \lambda_{2},\ \text{and}\ \lambda_{3}\) balance the relative contributions of the improvement reward, exploration incentive, and stagnation penalty.

Unlike conventional formulations that directly reward absolute fitness
values, the proposed reward is primarily driven by incremental fitness
improvement, encouraging the agent to favor actions that consistently
reduce the objective function, an effect analogous to a gradient-descent
process but without requiring explicit gradient information. The
diversity and stagnation terms further constrain the population to
maintain sufficient spread and escape local minima, together guiding the
DRL-DCO agent toward search dynamics that are both convergence-efficient
and robust for high-dimensional FEMU problems.

\hypertarget{ddpg-based-actorcritic-policy-gradient-scheme}{%
\subsubsection{\texorpdfstring{2.3.4 DDPG-based actor--critic policy
gradient scheme
}{2.3.4 DDPG-based actor--critic policy gradient scheme }}\label{ddpg-based-actorcritic-policy-gradient-scheme}}

Based on the reward function defined above, the objective of the DRL
agent (the DDPG-based actor--critic agent embedded within the DRL-DCO
algorithm) is to learn a control policy that maximizes the cumulative
reward over the optimization process. Formally, this can be expressed as
finding a policy \(\pi\) that maximizes the expected discounted return:

\[\operatorname{J}(\pi) = \mathbb{E}\left\lbrack \sum_{t = 0}^{T-1}\gamma^{t}r^{(t+1)} \right\rbrack\]

where \(\gamma \in (0,1)\) is a discount factor balancing immediate and
future rewards, and \(T\) is the number of generations in an episode.

This policy is learned through a DDPG-based actor--critic scheme,
constructed from two parametric neural networks: an actor and a critic.
The actor approximates a deterministic policy that maps the current
state \(\boldsymbol{s}^{(t)}\) to a continuous action
\(\boldsymbol{a}^{(t)}\), parameterized by
\(\boldsymbol{\theta}_{\pi}\), defined as
\(\boldsymbol{a}^{(t)} = \pi(\boldsymbol{s}^{(t)};\boldsymbol{\theta}_{\pi})\).
The critic approximates the action-value function
\(\operatorname{Q}^{\pi}(\boldsymbol{s}^{(t)},\boldsymbol{a}^{(t)})\),
i.e., the expected discounted return obtained by taking
\(\boldsymbol{a}^{(t)}\) in \(\boldsymbol{s}^{(t)}\) and following
\(\pi\) thereafter. It is parameterized by
\(\boldsymbol{\theta}_{\text{Q}}\), such that
\(\operatorname{Q}(\boldsymbol{s}^{(t)},\boldsymbol{a}^{(t)};\boldsymbol{\theta}_{\text{Q}}) \approx \operatorname{Q}^{\pi}(\boldsymbol{s}^{(t)},\boldsymbol{a}^{(t)})\).
Following the standard DDPG formulation, slowly updated target networks
with parameters \(\boldsymbol{\theta}_{\pi}'\) and
\(\boldsymbol{\theta}_{\text{Q}}'\) are used to compute the
temporal-difference (TD) target:

\[y^{(t)} = r^{(t+1)} + \gamma\left(1 - \mathbb{I}_{\text{term}}^{(t+1)}\right)\operatorname{Q}\left(\boldsymbol{s}^{(t+1)},\pi\left(\boldsymbol{s}^{(t+1)};\boldsymbol{\theta}_{\pi}'\right);\boldsymbol{\theta}_{\text{Q}}'\right)\]

where \(\mathbb{I}_{\text{term}}^{(t+1)} = \mathbb{I}\left(f_{\text{best}}^{(t+1)} < \varepsilon\right)\)
indicates that the episode terminates at generation \(t+1\), in which
case no bootstrapped term is used. The critic is trained by minimizing
the TD error with respect to \(\boldsymbol{\theta}_{\text{Q}}\), using
transitions sampled from the replay buffer \(\mathcal{D}\):

\[\mathcal{L}_{\text{Q}}\left(\boldsymbol{\theta}_{\text{Q}}\right) = \mathbb{E}_{\left(\boldsymbol{s}^{(t)},\boldsymbol{a}^{(t)},r^{(t+1)},\boldsymbol{s}^{(t+1)}\right)\sim\mathcal{D}}\left\lbrack\left(y^{(t)} - \operatorname{Q}\left(\boldsymbol{s}^{(t)},\boldsymbol{a}^{(t)};\boldsymbol{\theta}_{\text{Q}}\right)\right)^{2}\right\rbrack\]

The actor is updated by the deterministic policy gradient, i.e., by
minimizing the negative critic value of its own actions:

\[\mathcal{L}_{\pi}\left(\boldsymbol{\theta}_{\pi}\right) = -\mathbb{E}_{\boldsymbol{s}^{(t)}\sim\mathcal{D}}\left\lbrack\operatorname{Q}\left(\boldsymbol{s}^{(t)},\pi\left(\boldsymbol{s}^{(t)};\boldsymbol{\theta}_{\pi}\right);\boldsymbol{\theta}_{\text{Q}}\right)\right\rbrack\]

As discussed in Section 2.3.3, the reward is primarily governed by
incremental fitness improvement. Maximizing the expected return
therefore primarily encourages sustained objective reduction over
generations, analogous to a descent trajectory of the optimization
objective, while the diversity and stall terms preserve the flexibility
of stochastic, population-based exploration. Through this actor--critic
interaction, the agent learns to adaptively regulate the evolutionary
process based on observed optimization states, enabling efficient and
robust search in complex, high-dimensional model updating problems.

This formulation establishes the optimization process as a controlled
stochastic system in which population evolution is governed by the
DDPG-based actor--critic policy, mapping states to actions under reward
feedback. Building on this control framework, Section 2.4 specifies the
underlying evolutionary mechanisms being regulated, i.e., the dynamic
orchestration strategy that coordinates the DE and CMA-ES components
within each generation.

\hypertarget{dynamic-decma-es-orchestration}{%
\subsection{2.4 Dynamic DE--CMA-ES
orchestration}\label{dynamic-decma-es-orchestration}}

The formulation in Section 2.3 establishes the optimization process as a
controlled stochastic system, in which population evolution is governed
by a policy learned through the DDPG-based actor--critic framework.
Building on this framework, this section specifies how the learned
action variables are translated into adaptive regulation of the
evolutionary mechanisms within the DRL-DCO algorithm. Specifically, the
eight action variables introduced in Section 2.3.2, namely, \(F^{(t)}\),
\(CR^{(t)}\), \(\lambda^{(t)}\), \(LR_{\text{cov}}^{(t)}\),
\(\Delta\sigma^{(t)}\), \(p_{\text{res}}^{(t)}\), \(r_{\text{eli}}^{(t)}\), and
\(\beta^{(t)}\), jointly coordinate population composition, DE-based
variation, CMA-ES covariance-guided sampling, and intergenerational
preservation at each generation.

At generation \(t\), the effective population size \(N_{\text{eff}}^{(t)}\) is first adaptively regulated according to the population scaling factor \(\lambda^{(t)}\):

\[
N_{\text{eff}}^{(t)} = \mathrm{clip}\left( \lambda^{(t)}N_{\text{base}},\ N_{\min},\ N_{\max} \right)
\]

where \(N_{\text{base}}\) denotes the predefined base population size, while
\(N_{\min}\) and \(N_{\max}\) specify allowable population limits. A
fraction \(r_{\text{eli}}^{(t)}\) of the best-performing individuals is then
directly preserved through elite retention
\(N_{\text{eli}}^{(t)} = r_{\text{eli}}^{(t)}N_{\text{eff}}^{(t)}\), maintaining
intergenerational stability and prevents loss of high-quality candidate
solutions.

The remaining offspring are generated through dynamic orchestration
between the DE and the CMA-ES according to the orchestration ratio
\(\beta^{(t)}\), with offspring numbers allocated as:

% Version 1: Align the two rows of equations at "="
%\[
%\begin{aligned}
%N_{\text{DE}}^{(t)}  &= \beta^{(t)}\left( N_{\text{eff}}^{(t)} - N_{\text{eli}}^{(t)} \right),\\
%N_{\text{CMA}}^{(t)} &= \left(1 - \beta^{(t)}\right)\left( N_{\text{eff}}^{(t)} - N_{\text{eli}}^{(t)} \right)
%               = N_{\text{eff}}^{(t)} - N_{\text{eli}}^{(t)} - N_{\text{DE}}^{(t)}.
%\end{aligned}
%\]

% Version 2: Center the two rows of equations without alignment
\[\begin{gathered}
N_{\text{DE}}^{(t)} = \beta^{(t)}\left(N_{\text{eff}}^{(t)} - N_{\text{eli}}^{(t)}\right),\\
N_{\text{CMA}}^{(t)} = \left(1-\beta^{(t)}\right)\left(N_{\text{eff}}^{(t)} - N_{\text{eli}}^{(t)}\right)
= N_{\text{eff}}^{(t)} - N_{\text{eli}}^{(t)} - N_{\text{DE}}^{(t)}.
\end{gathered}\]

where \(\beta^{(t)} \in \lbrack 0,1\rbrack\) continuously regulates the
relative contribution between the DE-based directional exploration and
the CMA-ES covariance-guided exploitation.

\[\boldsymbol{v}_{i}^{(t)} = \boldsymbol{x}_{\text{r}_{1}}^{(t)} + F^{(t)}\left( \boldsymbol{x}_{\text{r}_{2}}^{(t)} - \boldsymbol{x}_{\text{r}_{3}}^{(t)} \right)\]

where \(\boldsymbol{x}_{\text{r}_{1}}^{(t)}\) \(,\) \(\boldsymbol{x}_{\text{r}_{2}}^{(t)}\),
and \(\boldsymbol{x}_{\text{r}_{3}}^{(t)}\) are randomly selected candidate
solutions, and \(\boldsymbol{v}_{i}^{(t)}\) is the mutated candidate. Trial
offspring are subsequently generated through crossover controlled by the
crossover probability \(CR^{(t)}\):

\[u_{i,j}^{(t)} = \left\{ \begin{matrix}
v_{i,j}^{(t)}, & \text{if~}{\text{rand}}(0,1) < CR^{(t)} \\
x_{i,j}^{(t)}, & \text{otherwise} \\
\end{matrix} \right.\ \]

where \(j\) denotes the parameter dimension. Larger values of
\(F^{(t)}\) and \(CR^{(t)}\) generally favor broader directional
exploration, while smaller values favor localized refinement. Further
details of the DE operators can be found in {[}38, 39{]}.

For the CMA-ES component, its offspring are sampled from an adaptive
multivariate Gaussian distribution:

\[\boldsymbol{x}_{i}^{(t)} \sim \mathcal{N}\left( \boldsymbol{m}^{(t)},\left( \sigma^{(t)} \right)^{2}\mathbf{C}^{(t)} \right)\]

where \(\boldsymbol{m}^{(t)}\), \(\mathbf{C}^{(t)}\), and \(\sigma^{(t)}\)
denote the search mean, covariance matrix, and step size, respectively.
Covariance adaptation is regulated by the action-controlled learning
factor \(LR_{\text{cov}}^{(t)}\), leading to:

\[\mathbf{C}^{(t + 1)} = \left( 1 - LR_{\text{cov}}^{(t)}c_{1} - LR_{\text{cov}}^{(t)}c_{\mu} \right)\mathbf{C}^{(t)} + LR_{\text{cov}}^{(t)}\Delta\mathbf{C}^{(t)}\]

where \(c_{1}\) and \(c_{\mu}\) denote standard CMA-ES learning
coefficients and \(\Delta\mathbf{C}^{(t)}\) is the covariance adaptation
term constructed from the weighted distribution of successful search
directions in the selected population, enabling the sampling
distribution to progressively align with promising directions of the
optimization landscape. Full update rules for \(\Delta\mathbf{C}^{(t)}\)
are provided in {[}10, 40{]}.

The step size \(\sigma^{(t)}\) is updated through the action-controlled
adjustment term \(\Delta\sigma^{(t)}\):

\[\sigma^{(t + 1)} = \sigma^{(t)}\exp\left( \Delta\sigma_{\text{CSA}}^{(t)} + \Delta\sigma^{(t)} \right)\]

where \(\Delta\sigma_{\text{CSA}}^{(t)}\) denotes the standard cumulative
step-size adaptation (CSA) update in the CMA-ES {[}10, 40{]}. Positive
values of \(\Delta\sigma^{(t)}\) encourage broader exploration, whereas
negative values promote local refinement, with the magnitude of
\(\Delta\sigma^{(t)}\) determining the strength of this adjustment.

A key feature of the proposed orchestration mechanism is that covariance
learning draws on top-performing candidates from the entire mixed
offspring population, rather than solely from CMA-ES-generated
solutions. Consequently, directional information discovered through DE
mutation also contributes to covariance adaptation, allowing the
Gaussian sampling distribution to align with promising search directions
identified across both operators.

To further improve robustness against premature convergence, stochastic
restart (introduced in Section 2.3.2) is applied to both DE- and
CMA-ES-generated offspring through the restart probability
\(p_{\text{res}}^{(t)}\): candidate generation is randomly reinitialized with
probability \(p_{\text{res}}^{(t)}\), enabling occasional exploration of
previously unvisited regions in highly multimodal search landscapes.

The proposed dynamic DE--CMA-ES orchestration differs fundamentally from
conventional hybrid EAs, which typically combine operators through
predefined rules, staged execution, or hard switching, imposing rigid
interactions that often require problem-specific manual tuning. By
contrast, the DRL-DCO algorithm formulates operator coordination as a
continuous control problem, where the embedded DRL agent adaptively
regulates population size, offspring allocation, and internal
evolutionary update rates according to the observed optimization state,
allowing the search process to transition smoothly among
exploration-dominant, exploitation-dominant, and mixed regimes.

Accordingly, for the high-dimensional FEMU and related optimization
problems, this adaptive orchestration offers two practical benefits: it
preserves the global exploration capability of the DE while utilizing
the covariance-learning mechanism of the CMA-ES for efficient local
refinement, and it avoids rigid hybridization rules, allowing the
framework to automatically adjust search behavior across optimization
stages while reducing the need for manual algorithm configuration.

\hypertarget{implementation-details-and-algorithm-workflow}{%
\subsection{2.5 Implementation details and algorithm
workflow}\label{implementation-details-and-algorithm-workflow}}

The DRL-DCO algorithm can operate in either of two distinct phases, each
capable of independently solving the optimization task, that together
govern its interaction between evolutionary optimization and DRL-based
policy learning. In the agent-training phase, the embedded DDPG-based
actor--critic agent operates simultaneously with population evolution,
as the evolutionary operators search for improved solutions, the actor
and critic networks are concurrently updated through deterministic
policy gradients and temporal-difference learning, respectively (Section
2.3.4). Alternatively, in the agent-inference phase, a previously
trained actor network instead generates generation-wise control actions
independently, without further parameter updates, critic evaluation, or
reward feedback, offering a more efficient deployment mode once training
has been completed.

To achieve stable adaptive regulation of the evolutionary process,
appropriate constraints are imposed on the action space. Although the
DRL-DCO framework enables autonomous adjustment of evolutionary
hyperparameters, unconstrained action outputs may lead to excessive
parameter variation, unstable population dynamics, or inefficient policy
exploration. Physically meaningful and computationally stable bounds are
therefore prescribed for the eight action variables generated by the
actor network, summarized in \textbf{Table 1}. These bounds follow two
principles: (i) maintaining sufficient flexibility for adaptive
exploration-exploitation regulation, and (ii) avoiding extreme control
behaviors that could compromise evolutionary stability or hinder
effective policy learning.

For the DE component, the mutation scaling factor \(F^{(t)}\) and
crossover probability \(CR^{(t)}\) are chosen based on effective DE
ranges to allow both local refinement and long-range exploration: the
lower bound of \(F^{(t)}\) ensures a slight movement of candidate
solutions to mitigate stagnation risk, while its upper bound of 2.0
permits strong oscillatory search behavior for escaping deep local
minima; for \(CR^{(t)}\), the lower bound prevents overly conservative
inheritance from the target vector, while the upper bound of 1.0 permits
full acceptance of the mutant vector when extensive exploration is
favorable.

For the CMA-ES component, the covariance learning rate \(LR_{\text{cov}}^{(t)}\)
controls the adaptation speed of the search distribution, while the
step-size adjustment factor \(\Delta\sigma^{(t)}\) enables adaptive
expansion or contraction of the search radius. Their bounded ranges
prevent abrupt distribution distortion, excessive exploration, or
premature convergence caused by inappropriate step-size evolution.

The remaining action variables regulate population-level and
hybridization behaviors. The population scaling factor \(\lambda^{(t)}\)
permits moderate variation in the effective population size, allowing
the algorithm to balance exploration and exploitation without incurring
excessive computational cost or risking population impoverishment at
small sizes. The restart probability \(p_{\text{res}}^{(t)}\) introduces
occasional stochastic reinitialization to facilitate escape from local
optima, while the elite preservation ratio \(r_{\text{eli}}^{(t)}\) balances
inheritance of high-quality solutions against population renewal,
avoiding both excessive elitism and search instability.

Finally, the orchestration ratio \(\beta^{(t)}\), primarily configured
within the bound {[}0.10, 0.90{]}, governs offspring allocation between
the CMA-ES and DE components. By excluding the extreme values of 0 and
1, this bound preserves continuous cooperative interaction between
covariance-guided exploitation and difference-vector-driven exploration,
allowing the DRL agent to dynamically prioritize one mechanism without
fully excluding the other. For strongly ill-conditioned, nonconvex
problems, \(\beta^{(t)}\) may additionally be constrained to a narrower,
more stable range of {[}0.30, 0.70{]}, ensuring a persistently balanced
exploration--exploitation tradeoff and more robust search behavior
throughout the optimization process.

In both phases, the DRL-DCO algorithm follows the same nested
episode-generation structure, in which the outer episode loop governs
long-term progress while the inner generation loop executes population
evolution. Within each episode, a maximum of 1000 generations is
employed, during which the algorithm sequentially performs dynamic
DE-CMA-ES orchestration and objective evaluation; only in the
agent-training phase are reward calculation, replay buffer storage, and
actor-critic parameter updating additionally performed, through which
the policy progressively improves its capability to regulate search
behavior according to the evolving optimization state.

To enhance training stability, the agent-training phase begins with a
warm-up stage prior to formal policy-guided optimization. Since
actor-critic learning relies on diverse experiences stored in the replay
buffer, insufficient initial samples may result in unstable gradient
estimation and biased policy updates. During the first five episodes,
populations are therefore independently and randomly initialized at the
start of each episode to generate diverse optimization trajectories
without inheriting previous search states. After this warm-up stage, the
best-performing optimization state obtained during initial exploration
is selected as the starting point for subsequent policy-guided
evolution, and training enters an inheritance-based mode in which
population information and the evolving policy progressively cooperate
to refine the solution.

The overall workflow of the DRL-DCO algorithm is summarized in the
flowchart in \textbf{Figure 1}, providing a visual overview of the
interaction between the DRL agent and the evolutionary operators, with
the corresponding algorithmic procedure summarized as pseudocode in
\textbf{Table 2}.

% ---------- Figure 1 (Section 2.5) ----------
\begin{figure}[H]
  \centering
  \includegraphics[width=0.9\linewidth]{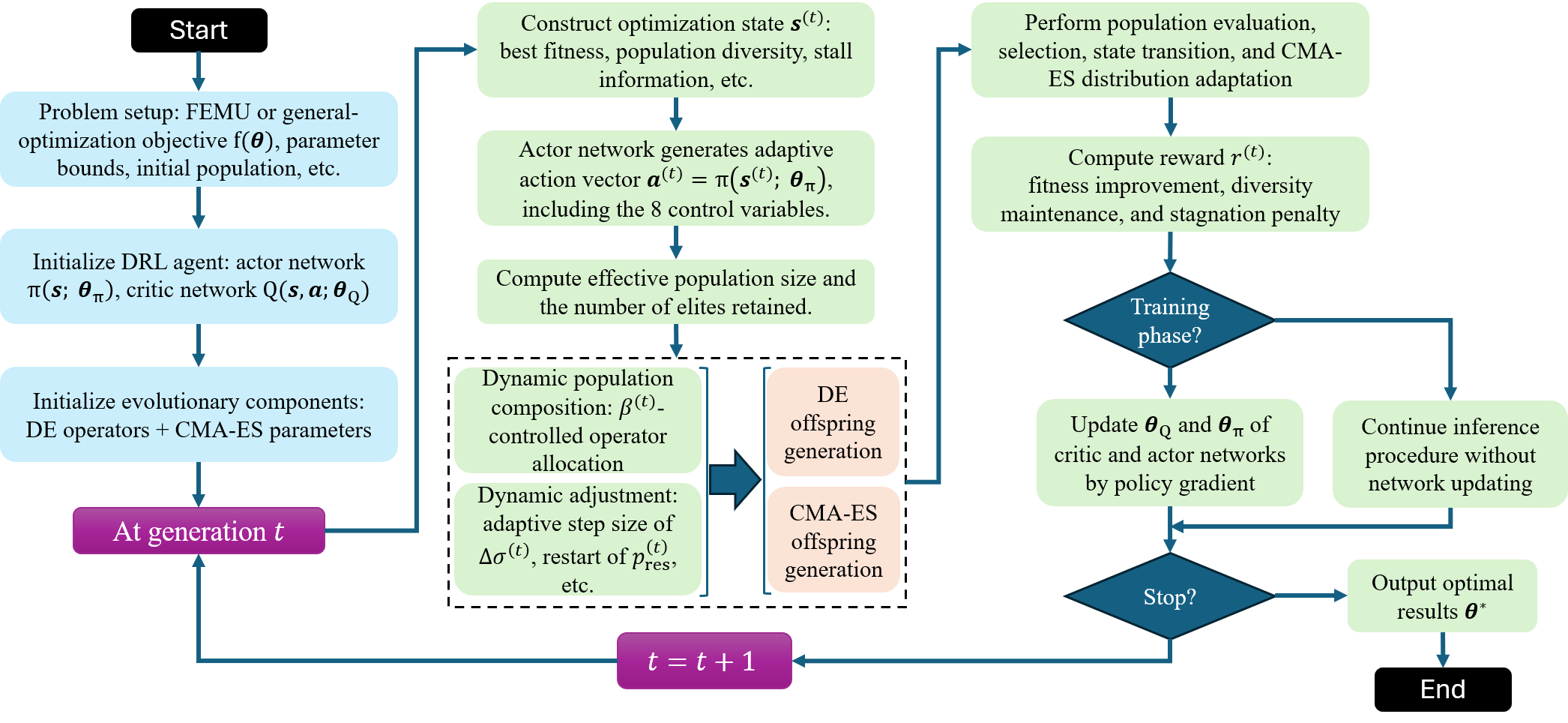}
  \caption{The flowchart of the proposed DRL-DCO algorithm.}
  \label{fig:flow}
\end{figure}

% ---------- Table 1 ----------
\begin{table}[H]
  \caption{Prescribed lower and upper bounds for the action variables in the DRL-DCO algorithm.}
  \label{tab:bounds}
  \centering\footnotesize\setlength{\tabcolsep}{4pt}
  \begin{tabularx}{\linewidth}{l*{4}{>{\centering\arraybackslash}X}}
    \toprule
    Variable & $F^{(t)}$ & $CR^{(t)}$ & $\lambda^{(t)}$ & $LR_{\mathrm{cov}}^{(t)}$ \\
    Range    & [0.01, 2.00] & [0.01, 1.00] & [0.75, 1.00] & [0.50, 1.50] \\
    \midrule
    Variable & $\Delta\sigma^{(t)}$ & $p_{\text{res}}^{(t)}$ & $r_{\text{eli}}^{(t)}$ & $\beta^{(t)}$ \\
    Range    & [$-$0.20, 0.20] & [0.00, 0.30] & [0.00, 0.30] & [0.10, 0.90]$^{\dagger}$ \\
    \bottomrule
  \end{tabularx}\\[2pt]
  {\scriptsize $^{\dagger}$ Narrowed to [0.30, 0.70] for strongly ill-conditioned, nonconvex problems.}
\end{table}

% ---------- Table 2 (pseudocode) ----------
\begin{table}[H]
  \caption{Pseudocode of the proposed DRL-DCO algorithm.}
  \label{tab:algo}
  \centering\footnotesize\setlength{\tabcolsep}{4pt}
  \begin{tabular}{@{}p{\dimexpr\linewidth-2\tabcolsep\relax}@{}}
    \toprule
    \textbf{Algorithm 1:} DRL-governed dynamic DE--CMA-ES Orchestration (DRL-DCO) \\
    \midrule
    \textbf{Input:} FEMU or general single-objective objective function $\operatorname{f}(\boldsymbol{\theta})$, search bounds, population size, maximum episodes $N_{\text{epi}}$, maximum generations per episode $N_{\text{gen}}$. \\
    \textbf{Output:} The optimal solution vector $\boldsymbol{\theta}^{*}$. \\
    \midrule
    1:\; Initialize DE operators, CMA-ES parameters, actor $\pi(\boldsymbol{s};\boldsymbol{\theta}_\pi)$, critic $\operatorname{Q}(\boldsymbol{s},\boldsymbol{a};\boldsymbol{\theta}_\text{Q})$, and replay memory. \\
    2:\; \textbf{for} episode $e = 1,\ldots,N_{\text{epi}}$ \textbf{do} \\
    3:\; \hspace{1.2em}Initialize or inherit population $\boldsymbol{\chi}^{(0)}$ according to the warm-up/training strategy. \\
    4:\; \hspace{1.2em}Construct initial state $\boldsymbol{s}^{(0)}$ from $\boldsymbol{\chi}^{(0)}$. \\
    5:\; \hspace{1.2em}\textbf{for} generation $t = 0,\ldots,N_{\text{gen}}-1$ \textbf{do} \\
    6:\; \hspace{2.4em}Generate action $\boldsymbol{a}^{(t)} = \pi(\boldsymbol{s}^{(t)};\boldsymbol{\theta}_\pi) = [F,CR,\lambda,LR_{\text{cov}},\Delta\sigma,p_{\text{res}},r_{\text{eli}},\beta]^{(t)}$. \\
    7:\; \hspace{2.4em}Orchestrate: generate offspring with DE and CMA-ES, allocated through $\beta^{(t)}$. \\
    8:\; \hspace{2.4em}Evaluate offspring using $\operatorname{f}(\boldsymbol{\theta})$. \\
    9:\; \hspace{2.4em}Update population to $\boldsymbol{\chi}^{(t+1)}$, perform CMA-ES adaptation, and construct $\boldsymbol{s}^{(t+1)}$. \\
    10:\; \hspace{2.4em}\textbf{if} agent-training phase \textbf{then} \\
    11:\; \hspace{3.6em}Compute reward $r^{(t+1)}$ from the transition $\boldsymbol{s}^{(t)} \to \boldsymbol{s}^{(t+1)}$. \\
    12:\; \hspace{3.6em}Store $(\boldsymbol{s}^{(t)},\boldsymbol{a}^{(t)},r^{(t+1)},\boldsymbol{s}^{(t+1)})$ in replay memory. \\
    13:\; \hspace{3.6em}Update $\boldsymbol{\theta}_\pi$ and $\boldsymbol{\theta}_\text{Q}$ by the deterministic policy gradient method. \\
    14:\; \hspace{2.4em}\textbf{else} inference only (no reward, no network updates). \\
    15:\; \hspace{2.4em}\textbf{if} $f_{\text{best}}^{(t+1)} < \varepsilon$ \textbf{then} terminate the episode early. \\
    16:\; \hspace{1.2em}\textbf{end for}; record the best solution of the current episode. \\
    17:\; \textbf{end for}; return $\boldsymbol{\theta}^{*}$. \\
    \bottomrule
  \end{tabular}
\end{table}

\hypertarget{results}{%
\section{3. Results}\label{results}}

\hypertarget{testing-on-benchmark-optimization-problems}{%
\subsection{3.1 Testing on benchmark optimization
problems}\label{testing-on-benchmark-optimization-problems}}

\hypertarget{benchmark-functions-and-baseline-algorithms}{%
\subsubsection{3.1.1 Benchmark functions and baseline
algorithms}\label{benchmark-functions-and-baseline-algorithms}}

To evaluate the effectiveness and generalization capability of the
proposed DRL-DCO algorithm prior to its application to FEMU, a set of
well-established benchmark optimization functions was first employed as
a systematic validation testbed. Five representative single-objective
benchmark functions were adopted: the Ackley, Griewank, Rastrigin,
Rosenbrock, and Schwefel functions {[}41{]}. These functions are
standard test cases in the optimization literature because they
collectively span a diverse and complementary range of challenging
landscape properties, including multimodality, non-convexity,
inter-variable coupling, ill-conditioning, and deceptive optima,
enabling a rigorous assessment of global search capability, local
refinement precision, and robustness against premature convergence.

Briefly, the Ackley function combines a flat outer region with a sharply
structured, multi-optima central basin; the Griewank function exhibits
widespread low-amplitude oscillations with strong inter-variable
coupling that intensifies with dimensionality; the Rastrigin function
presents a highly multimodal landscape with densely and regularly
distributed local minima; the Rosenbrock function forms a narrow,
ill-conditioned curved valley that is comparatively easy to locate but
difficult to traverse toward its unique global optimum; and the Schwefel
function is strongly deceptive, with its global optimum located far from
the majority of local optima. Together, these properties test
complementary aspects of exploration, exploitation, and convergence
robustness, consistent with the practical implications of the No Free
Lunch (NFL) theorem {[}42{]}, which states that no single algorithm can
universally dominate across all problem classes. This motivates the
adaptive hyperparameter regulation and dynamic operator orchestration
central to the proposed DRL-DCO algorithm, whose embedded DDPG-based
actor-critic agent continuously modulates search behavior according to
the observed optimization state, aiming to enhance robustness and
flexibility across this diverse benchmark suite.

All five benchmark functions were evaluated at three problem
dimensionalities, \(N_{\text{d}}\) = 20, 30, and 50. As \(N_{\text{d}}\) = 50
represents the most challenging and practically informative setting,
results for this dimensionality are reported in the main text; results
at \(N_{\text{d}}\) = 20 and 30 follow consistent trends and are omitted for
brevity. Search domains follow the canonical standards established in
the optimization literature {[}41, 42, 43{]} and are summarized,
together with the global optimum value and location for each function,
in \textbf{Table 3}.

Collectively, these benchmark functions present complementary and
mutually reinforcing optimization challenges, including multimodality,
non-convexity, inter-variable coupling, ill-conditioning, deceptive
local optima, and exploration--exploitation imbalance. Consequently,
achieving consistently strong performance across all benchmark functions
at increasing dimensionality is inherently difficult, and reflects the
practical implications of the NFL theorem, which establishes that no
single optimization algorithm can universally outperform all others
across all problem classes. Motivated by this challenge, the proposed
DRL-DCO algorithm incorporates adaptive hyperparameter regulation and
dynamic operator orchestration through an embedded DDPG-based
actor--critic agent, which outputs continuous generation-wise control
actions to continuously modulate search behavior in response to the
observed optimization state. This adaptive mechanism is designed to
enhance robustness and flexibility across the diverse optimization
landscapes represented by the benchmark suite.

For comparative evaluation, five baseline algorithms were considered: 1)
JADE, an adaptive DE variant employing the current-to-pbest mutation
strategy with adaptive parameter sampling and an optional external
archive {[}38{]}; 2) LSHADE, the CEC 2014 competition-winning algorithm
extending JADE's success-history-based parameter adaptation with a
linear population size reduction mechanism {[}39{]}; 3) LSHADE-SPACMA, a
well-established self-adaptive hybrid algorithm integrating LSHADE and
CMA-ES mechanisms {[}44{]}; 4) DRL-DE; and 5) DRL-CMA-ES. Here, the
DRL-DE and DRL-CMA-ES serve as 2 ablation variants of the proposed
DRL-DCO framework, obtained by fixing the orchestration ratio
\(\beta^{(t)}\) to 1 and 0, respectively, such that the DDPG-based agent
regulates only the DE or CMA-ES component while the overall RL control
architecture remains identical to that of DRL-DCO. This design enables
rigorous assessment of the contribution of the dynamic DE--CMA-ES
orchestration mechanism beyond single-operator DRL-governed
optimization. Together, the five baselines form a progressive
algorithmic hierarchy, from modern adaptive DE (JADE), to
competition-level adaptive DE with population reduction (LSHADE), to a
state-of-the-art non-DRL hybrid (LSHADE-SPACMA), to the two
DRL-augmented single-operator ablations, enabling systematic attribution
of performance gains to each successive design advancement. The initial
population size was uniformly set to 100 for all algorithms.
Hyperparameters for the JADE, LSHADE, and LSHADE-SPACMA followed the
recommended configurations in their respective original references
{[}38, 39, 44{]}, while the hyperparameters of the DRL-DCO follow the
specifications in \textbf{Table 1}.

The primary performance metric is the total number of generations
required to reach full convergence to the global optimum, declared when
the absolute deviation between the obtained and true global optimum
falls below \(\varepsilon\) = $1\times10^{-8}$. For the three DRL-based algorithms (DRL-DCO, DRL-DE, and DRL-CMA-ES), the total generation count is the cumulative number of generations executed across episodes (each capped at 1,000 generations) until the convergence threshold is reached;
for the three conventional EA baselines, which operate without an
episodic structure, the total generation count is recorded directly. A
uniform budget of 40,000 total generations was imposed across all
algorithms (equivalent to 40 episodes of 1,000 generations per episode
for the DRL-based methods), beyond which optimization was terminated
regardless of convergence status.

% ---------- Table 3 ----------
\begin{table}[H]
  \caption{Standard search domains, global optima, and optimal points of the five benchmark optimization functions adopted for algorithm validation.}
  \label{tab:bench}
  \centering\footnotesize\setlength{\tabcolsep}{6pt}
  \begin{tabular}{@{}lcc@{}}
    \toprule
    Function & Standard search domain & Global optimum \\
    \midrule
    Ackley     & $[-32.768,\,32.768]^{N_\text{d}}$ & $\operatorname{f}(\boldsymbol{x}^*)=0$ at $\boldsymbol{x}^*=\mathbf{0}$ \\
    Griewank   & $[-600,\,600]^{N_\text{d}}$       & $\operatorname{f}(\boldsymbol{x}^*)=0$ at $\boldsymbol{x}^*=\mathbf{0}$ \\
    Rastrigin  & $[-5.12,\,5.12]^{N_\text{d}}$     & $\operatorname{f}(\boldsymbol{x}^*)=0$ at $\boldsymbol{x}^*=\mathbf{0}$ \\
    Rosenbrock & $[-30,\,30]^{N_\text{d}}$         & $\operatorname{f}(\boldsymbol{x}^*)=0$ at $\boldsymbol{x}^*=\mathbf{1}$ \\
    Schwefel   & $[-500,\,500]^{N_\text{d}}$       & $\operatorname{f}(\boldsymbol{x}^*)=0$ at $\boldsymbol{x}^*=\mathbf{420.9687}$ \\
    \bottomrule
  \end{tabular}
\end{table}

\hypertarget{benchmark-analysis-of-agent-training-phase}{%
\subsubsection{\texorpdfstring{3.1.2 Benchmark analysis of
agent-training phase
}{3.1.2 Benchmark analysis of agent-training phase }}\label{benchmark-analysis-of-agent-training-phase}}

As described in Section 2.5, the three DRL-based algorithms, including
DRL-DCO and its two ablation variants of DRL-DE and DRL-CMA-ES, are
first evaluated under the agent-training phase, where the DDPG-based
actor--critic agent learns adaptive control policies through interaction
with the optimization environment. During this phase, evolutionary
optimization and policy learning are performed simultaneously, allowing
the agent to progressively improve its regulation of the search process.
The corresponding agent-inference performance after policy training is
further investigated in Section 3.1.3.

All three DRL-based algorithms incorporate a five-episode warm-up period
(5000 generations) at the start of each optimization run, during which
the population is independently reinitialized at each episode to
accumulate sufficient replay buffer experience for stable policy
learning (Section 2.5). This warm-up mechanism delays convergence onset
and moderately penalizes early-stage performance, but in doing so
ensures more reliable actor-critic training, making the resulting
comparison more stringent and the reported convergence performance more
conservative and credible.

The convergence results of all six algorithms on the five benchmark
functions at \(N_{\text{d}} = 50\) are summarized in \textbf{Table 4}. For the
two widely recognized challenging cases, Rastrigin and Rosenbrock,
selected as representative examples, the best-fitness convergence
histories and effective population distribution are further shown in
\textbf{Figure 2}, while the corresponding evolution of key action
variables is shown in \textbf{Figure 3}. Several key observations are
drawn from these results.

1) The developed DRL-DCO algorithm achieves consistent, well-rounded
convergence across all five benchmark cases, with its generation cost,
function evaluation count, and wall-clock time consistently ranking at
or above the median among the six algorithms. This balanced performance,
spanning landscapes as different as the dense multimodality of Rastrigin
and Schwefel, and the ill-conditioned valley geometry of Rosenbrock,
demonstrates that the learned orchestration policy generalizes across
fundamentally different problem structures without sacrificing
performance on any single class to gain on another.

2) Examining Rastrigin and Schwefel results in detail reveals a sharp
divergence between CMA-ES-dominant and DE-based strategies. Among the
non-DRL baselines, LSHADE-SPACMA substantially outperforms the purely
DE-based JADE and LSHADE on Rastrigin, confirming that hybridizing DE
and CMA-ES benefits dense multimodal landscapes. Among the DRL-based
algorithms, however, DRL-CMA-ES (\(\beta^{(t)}\) \(= 0\)) fails entirely
on both functions, converging prematurely to suboptimal basins, which is
consistent with CMA-ES's known tendency toward single-basin
distributional collapse on regularly multimodal landscapes {[}45, 46{]}.
The results here again demonstrate that stochastic restarts alone,
without an accompanying population size increase, are insufficient to
reliably escape these deceptive local basins {[}46{]}. In contrast, the
DE-based algorithms of DRL-DE, JADE, and LSHADE all eventually converge
on these functions, since difference-vector mutation is recomputed fresh
from the population's spread at every generation and continues
generating exploratory candidates as long as diversity is preserved
{[}9, 39{]}. This robustness comes at a cost, however, since exploring
multimodal landscapes largely by attrition rather than
curvature-informed search is why JADE and LSHADE require substantially
more generations than LSHADE-SPACMA, and why DRL-DE requires more
generations than DRL-DCO, to reach the global optimum on Rastrigin.

3) The Rosenbrock results reveal an opposite and more nuanced pattern,
where the classical advantage of DE-based strategies proves conditional
rather than intrinsic. Among the non-DRL baselines, JADE and LSHADE
outperform LSHADE-SPACMA, reflecting DE's advantage on
Rosenbrock\textquotesingle s continuously curving valley: lacking
directional memory, difference-vector mutation can pivot immediately to
track the valley's rotation, whereas CMA-ES's evolution-path-based
covariance accumulation introduces a momentum-like lag that causes the
covariance matrix to perpetually trail the valley's curvature {[}47{]}.
Among the DRL-based algorithms, this relationship is strikingly
reversed: DRL-DE fails entirely, exhibiting prolonged stagnation with
only sporadic, restart-driven improvement, while DRL-CMA-ES reaches the
optimum through slow but steady progression. This reversal is
attributable to the absence, in the DRL-DE implementation, of the
sophisticated adaptive mechanisms present in LSHADE (success-history
parameter memory, external archive, and linear population size
reduction). Without these mechanisms providing systematic population
management and parameter diversity, the structural limitation of DE, the
coordinate-axis bias of binomial crossover, which prevents it from
efficiently producing the arbitrarily oriented, anisotropic refinement
steps demanded near the optimum of Rosenbrock, becomes the dominant
factor. This confirms that the advantage of DE on Rosenbrock depends on
the sophistication of its accompanying adaptive mechanisms, rather than
being an unconditional property of the difference-vector mutation
operator itself.

4) Tracking the elite and orchestration ratio histories of DRL-DCO in
Figure 3 further illustrates how the learned policy evolves during
training. The warm-up period provides the agent with an initial,
unconstrained exploration space in which to accumulate first-hand
experience and establish a preliminary search direction. Once training
stabilizes, the agent settles into a narrower elite ratio, and adjusts its orchestration ratio according to the problem
at hand. On the nonconvex Rastrigin and Rosenbrock functions
specifically, the stabilized policy favors the DE component while
retaining a partial CMA-ES contribution, preserving the directional
flexibility needed to escape local minima without fully discarding the
covariance-based momentum that supports faster convergence. On the
comparatively easier Griewank function, by contrast, this adaptive
orchestration proves largely unnecessary: both DRL-CMA-ES and DRL-DCO
already reach the global optimum within the first warm-up episode (1000
generations), while the actor-critic policy is still in its early,
largely untrained stage, indicating that the underlying evolutionary
operators alone are sufficient for this landscape.

5) Overall, the DRL-DE and DRL-CMA-ES ablations, obtained by fixing
\(\beta^{(t)}\) to 1 and 0 respectively, show markedly inferior and less
reliable convergence, particularly on Rastrigin, Rosenbrock, and
Schwefel, where at least one of the two exhausts the maximum generation
budget without reaching the convergence threshold. This demonstrates
that the orchestration ratio is not a peripheral hyperparameter but a
structurally essential mechanism, since committing to a single operator
leaves the algorithm unable to compensate for that
operator\textquotesingle s inherent limitations on mismatched
landscapes, regardless of how well its remaining hyperparameters are
tuned. On Rastrigin, the generation-wise \(\beta^{(t)}\) trajectory
shows the agent learning to alternate between CMA-ES-dominant phases for
rapid descent into candidate basins and DE-dominant phases for diversity
restoration, effectively reproducing the role of population-restart
strategies such as IPOP-CMA-ES {[}46{]} through continuous reweighting
rather than discrete reinitialization. On Rosenbrock, the agent learns
to engage CMA-ES for anisotropic local refinement while periodically
invoking DE-driven population relocation to escape curvature-tracking
stagnation. These emergent behaviors confirm that the learned
orchestration policy captures complementary operator strengths in a
problem-adaptive, trajectory-aware manner that neither fixed
hybridization (LSHADE-SPACMA) nor single-operator DRL control (DRL-DE,
DRL-CMA-ES) can replicate.

Taken together, these results reflect a consistent, theoretically
grounded pattern, echoing the NFL theorem's prediction that no single
operator or fixed hybridization strategy achieves uniformly strong
performance across all landscape types. LSHADE-SPACMA, the strongest
non-DRL baseline, excels on Rastrigin but lags on Rosenbrock, while JADE
and LSHADE show the opposite tendency and perform less effectively on Ackley and
Griewank at \(N_{\text{d}} = 50\). Similarly, DRL-DE and DRL-CMA-ES each
succeed where the other fails. DRL-DCO, through its learned
generation-wise orchestration of both operators in response to the
observed optimization state, achieves consistently competitive
performance across all five functions, demonstrating that dynamic,
state-aware operator coordination is not merely beneficial but necessary
for robust high-dimensional optimization across diverse problem classes.

% ---------- Figure 2 ----------
\begin{figure}[H]
  \centering
  \includegraphics[width=0.75\linewidth]{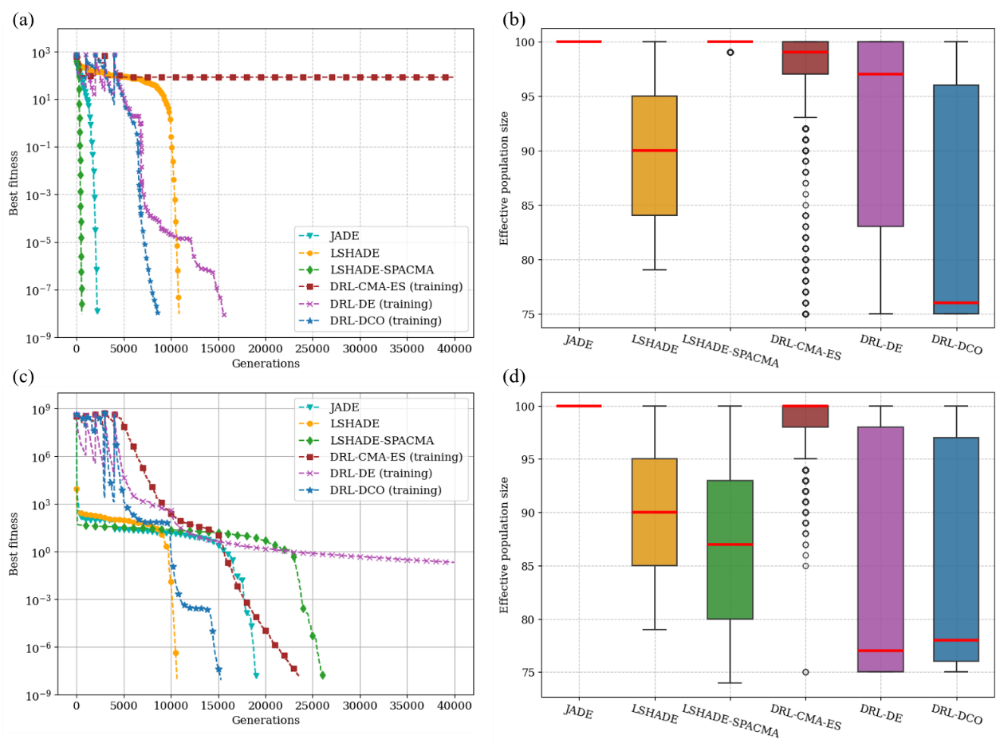}
  \caption{Convergence histories of best fitness and effective population distribution of the five baseline algorithms and the DRL-DCO algorithm (agent-training phase) on the Rastrigin and Rosenbrock functions at dimensionality $N_\text{d} = 50$. \textbf{(a)} Convergence history on Rastrigin; \textbf{(b)} effective population size distribution on Rastrigin; \textbf{(c)} convergence history on Rosenbrock; \textbf{(d)} effective population size distribution on Rosenbrock.}
\end{figure}

% ---------- Figure 3 ----------
\begin{figure}[H]
  \centering
  \includegraphics[width=0.8\linewidth]{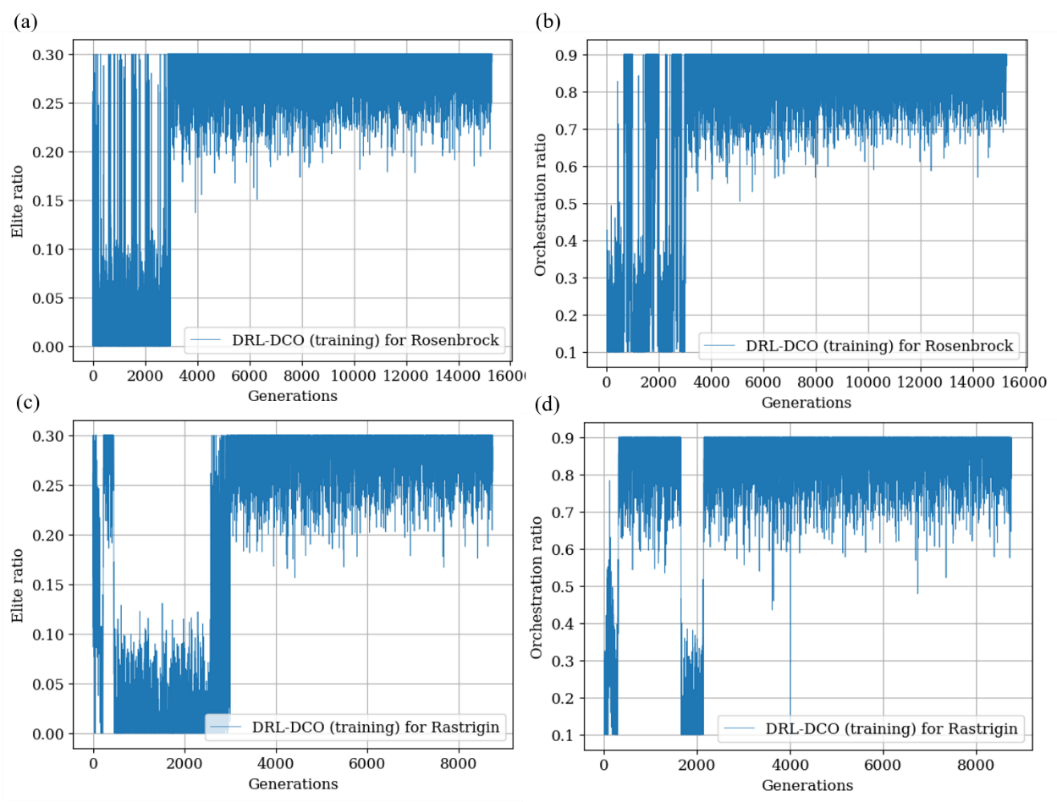}
  \caption{Evolution of the elite ratio and orchestration ratio in the DRL-DCO algorithm (agent-training phase) on the Rastrigin and Rosenbrock functions at dimensionality $N_\text{d} = 50$. \textbf{(a)} Elite ratio evolution on Rosenbrock; \textbf{(b)} orchestration ratio evolution on Rosenbrock; \textbf{(c)} elite ratio evolution on Rastrigin; \textbf{(d)} orchestration ratio evolution on Rastrigin.}
\end{figure}

% ---------- Table 4 ----------
\begin{table}[H]
  \caption{Convergence performance of the five baseline algorithms and the developed DRL-DCO algorithm (agent-training phase) on the five benchmark optimization functions at dimensionality $N_\text{d} = 50$. All results are averages over 10 independent runs, with standard deviations in parentheses. ``Failure'' indicates that none of the 10 independent runs achieved the convergence threshold $\varepsilon = 1\times10^{-8}$ within the maximum budget of 40,000 total generations, in which case no standard deviation is reported.}
  \label{tab:train}
  \centering\footnotesize\setlength{\tabcolsep}{4pt}
  \begin{tabular}{@{}llrrr@{}}
    \toprule
    Function & Algorithm & Avg total generations & Avg total function evaluations & Approx. time (s) \\
    \midrule
    \multirow{6}{*}{Ackley}
     & JADE          & 40,000 (failure) & 4,000,000 (failure) & 890--990 \\
     & LSHADE        & 573 (118)        & 56,988 (11,756)     & 14--18 \\
     & LSHADE-SPACMA & 386 (72)         & 38,592 (7,143)      & 10--13 \\
     & DRL-CMA-ES    & 18,194 (1,023)   & 1,268,258 (99,143)  & 230--300 \\
     & DRL-DE        & 15,956 (1,312)   & 1,139,602 (105,672) & 310--390 \\
     & DRL-DCO       & 5,923 (556)      & 514,622 (47,443)    & 85--130 \\
    \midrule
    \multirow{6}{*}{Griewank}
     & JADE          & 40,000 (failure) & 4,000,000 (failure) & 890--990 \\
     & LSHADE        & 406 (72)         & 40,277 (7,143)      & 10--12 \\
     & LSHADE-SPACMA & 312 (55)         & 31,188 (5,467)      & 8--10 \\
     & DRL-CMA-ES    & 625 (183)        & 51,606 (15,143)     & 7--12 \\
     & DRL-DE        & 16,042 (1,534)   & 1,477,821 (151,926) & 280--420 \\
     & DRL-DCO       & 578 (198)        & 47,624 (16,434)     & 6--14 \\
    \midrule
    \multirow{6}{*}{Rastrigin}
     & JADE          & 2,207 (438)      & 220,700 (43,800)    & 45--60 \\
     & LSHADE        & 10,869 (1,189)   & 973,506 (90,812)    & 240--295 \\
     & LSHADE-SPACMA & 571 (102)        & 57,030 (10,187)     & 14--19 \\
     & DRL-CMA-ES    & 40,000 (failure) & 3,212,779 (failure) & 550--670 \\
     & DRL-DE        & 15,611 (2,923)   & 1,041,326 (208,765) & 275--415 \\
     & DRL-DCO       & 8,753 (1,272)    & 634,950 (95,563)    & 105--165 \\
    \midrule
    \multirow{6}{*}{Rosenbrock}
     & JADE          & 19,044 (2,712)   & 1,904,400 (271,200) & 415--480 \\
     & LSHADE        & 10,645 (1,312)   & 953,693 (107,434)   & 235--290 \\
     & LSHADE-SPACMA & 26,083 (6,204)   & 2,268,155 (506,615) & 590--720 \\
     & DRL-CMA-ES    & 23,657 (5,023)   & 2,007,623 (455,734) & 310--430 \\
     & DRL-DE        & 40,000 (failure) & 2,854,277 (failure) & 780--980 \\
     & DRL-DCO       & 15,301 (2,153)   & 979,228 (141,312)   & 165--250 \\
    \midrule
    \multirow{6}{*}{Schwefel}
     & JADE          & 2,572 (358)      & 257,200 (35,800)    & 55--70 \\
     & LSHADE        & 9,787 (912)      & 886,767 (85,678)    & 220--270 \\
     & LSHADE-SPACMA & 16,253 (2,823)  & 1,493,260 (243,256) & 385--475 \\
     & DRL-CMA-ES    & 40,000 (failure) & 3,018,872 (failure) & 550--670 \\
     & DRL-DE        & 18,872 (2,878)   & 1,486,744 (220,234) & 335--490 \\
     & DRL-DCO       & 10,219 (1,023)   & 786,905 (75,618)    & 130--200 \\
    \bottomrule
  \end{tabular}
\end{table}

\hypertarget{benchmark-analysis-of-agent-inference-phase}{%
\subsubsection{3.1.3 Benchmark analysis of agent-inference
phase}\label{benchmark-analysis-of-agent-inference-phase}}

Following completion of the agent-training phase, the trained DRL-DCO
algorithm was further evaluated under its agent-inference phase, in
which the actor network's output actions directly determine
generation-wise control decisions without further weight updates to
either the actor or critic. In this mode, the trained agent autonomously
governs all key algorithmic configurations, including the DE and CMA-ES
operator parameters, effective population size, resource allocation
between the two operators, and restart behavior, relying entirely on its
internalized policy to navigate the search landscape. As this
agent-inference capability tests generalization of the orchestration policy specifically, and the non-DRL baselines possess no such learned policy, while DRL-DE and DRL-CMA-ES fix $\beta^{(t)}$ and thus retain no such orchestration mechanism, only DRL-DCO is evaluated here.

The agent-inference phase represents a qualitatively distinct mode of
operation from both the adaptive EA baselines and the agent-training
phase itself. The inference phase relies solely on the actor
network\textquotesingle s forward pass to determine each control action,
eliminating the critic evaluation and temporal-difference weight updates
required during training. This yields a notably lower
per-generation computational cost while retaining the full adaptive
intelligence encoded in the trained policy, allowing the inference-phase
DRL-DCO to independently locate the global optimum across problem
instances without requiring online supervision or feedback-driven
adjustment.

Representative convergence histories of the agent-inference phase,
including best-fitness trajectories, effective population size, and
orchestration ratio for the Rastrigin and Rosenbrock functions, are
presented in \textbf{Figures 4} and \textbf{5}, respectively. The
complete agent-inference results across all five benchmark functions,
averaged over 10 independent runs with randomized population
initialization, are summarized in \textbf{Table 5}. Several key
observations follow.

1) The agent-inference phase consistently converges in fewer total generations than the corresponding agent-training phase
across all benchmark functions, confirming that eliminating the warm-up
period directly translates into earlier effective search onset. This
validates the core premise of the two-phase design, the agent-training
phase functions as a one-time investment in policy learning, after which
the trained agent can be deployed with immediate, unsupervised
effectiveness, demonstrating strong generalization and intelligence from
a single 40-episode training realization to diverse problem instances
with randomized initializations.

2) Although this generation reduction is consistent across all
functions, its magnitude is in most cases smaller than the 5,000
generations corresponding to the full warm-up period, with reductions of
approximately 3,000 generations observed for Rastrigin and approximately 900 generations for Rosenbrock. This indicates that the inference phase
does not simply recover the warm-up overhead and nothing more, since the
absence of online critic feedback introduces a degree of action noise
relative to the fully supervised training setting, meaning the trained
policy, while generally effective, does not always produce identically
optimal generation-wise decisions across independent runs. The moderate
standard deviations in total generation and function evaluation counts,
notably lower than their training-phase counterparts, confirm that this
variability remains well controlled and that the inference phase is
stable across all benchmark functions.

3) The agent-inference phase is consistently the fastest per generation
among all evaluated configurations, including the agent-training phase
and the three adaptive EA baselines, though the source of this advantage
differs across comparisons. Relative to the training phase, the gain is
structural and substantial. Removing the critic forward pass,
temporal-difference target computation, and actor-critic weight updates
leaves only a single actor forward pass and standard population
evaluation, eliminating the entire RL update overhead that the training
phase must otherwise carry at every generation. Relative to the adaptive
EA baselines, this efficiency advantage is more modest, mainly stemming
from the actor's forward pass being lighter than the adaptive
bookkeeping, such as success-history memory and archive maintenance,
that JADE, LSHADE, and LSHADE-SPACMA carry out every generation.

To summarize, these results demonstrate that DRL-DCO achieves a
meaningful and practically useful separation between its learning and
deployment costs. The trained policy generalizes reliably across
independently initialized runs on all five benchmark functions,
producing convergence performance that is both faster per generation
than the training phase and competitive with, or superior to, the best
non-DRL baselines in total generation cost, all without requiring online
adaptation or feedback during deployment. This confirms that the learned
orchestration paradigm offers genuine operational advantages beyond
training, and that the intelligence acquired during agent-training
translates into a self-contained, supervision-free optimization
capability.

% ---------- Table 5 ----------
\begin{table}[H]
  \caption{Convergence performance of the developed DRL-DCO algorithm (agent-inference phase) on the five benchmark optimization functions at dimensionality $N_\text{d} = 50$, obtained from one round of the agent-training phase (40 episodes). All results are averages over 10 independent runs, with standard deviations in parentheses.}
  \label{tab:infer}
  \centering\footnotesize\setlength{\tabcolsep}{6pt}
  \begin{tabular}{@{}lrrr@{}}
    \toprule
    Function & Avg total generations & Avg total function evaluations & Approx. time (s) \\
    \midrule
    Ackley     & 5,176 (202)  & 406,656 (16,195)   & 32--35 \\
    Griewank   & 556 (63)     & 42,571 (3,014)     & 4--6 \\
    Rastrigin  & 5,510 (217)  & 386,286 (17,276)   & 33--36 \\
    Rosenbrock & 14,331 (498) & 1,003,834 (45,277) & 90--96 \\
    Schwefel   & 8,989 (379)  & 677,712 (31,168)   & 52--57 \\
    \bottomrule
  \end{tabular}
\end{table}

% ---------- Figure 4 ----------
\begin{figure}[H]
  \centering
  \includegraphics[width=0.8\linewidth]{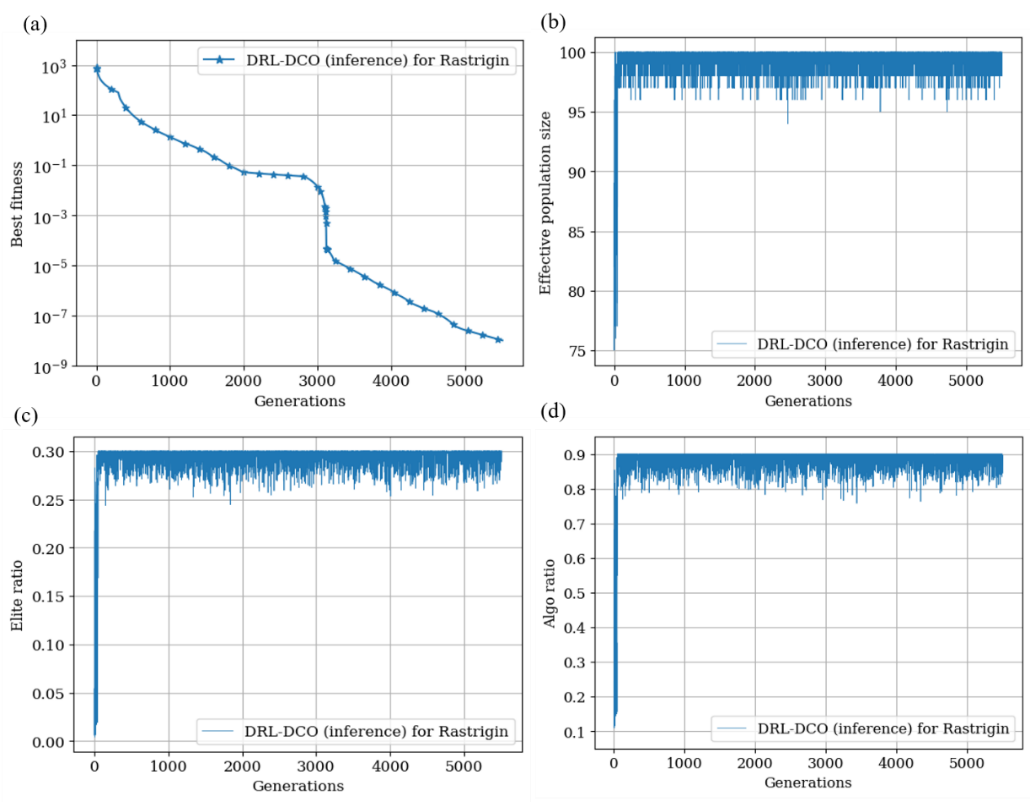}
  \caption{Convergence performance of the developed DRL-DCO algorithm during the agent-inference phase on the Rastrigin function at dimensionality $N_\text{d} = 50$. \textbf{(a)} Convergence history; \textbf{(b)} effective population size evolution; \textbf{(c)} elite ratio evolution; \textbf{(d)} orchestration ratio evolution.}
\end{figure}

% ---------- Figure 5 ----------
\begin{figure}[H]
  \centering
  \includegraphics[width=0.8\linewidth]{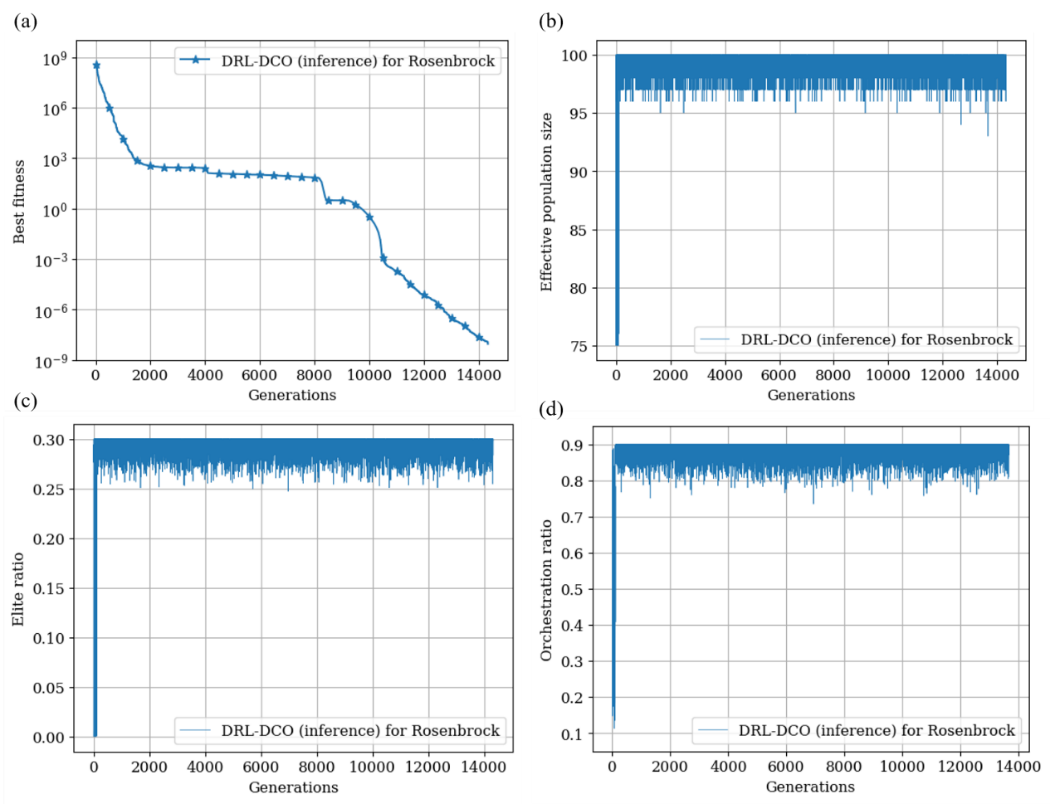}
  \caption{Convergence performance of the developed DRL-DCO algorithm during the agent-inference phase on the Rosenbrock function at dimensionality $N_\text{d} = 50$. \textbf{(a)} Convergence history; \textbf{(b)} effective population size evolution; \textbf{(c)} elite ratio evolution; \textbf{(d)} orchestration ratio evolution.}
\end{figure}

\hypertarget{application-to-femu-iasc-asce-benchmark-validation}{%
\subsection{3.2 Application to FEMU: IASC-ASCE benchmark
validation}\label{application-to-femu-iasc-asce-benchmark-validation}}

\hypertarget{essential-configurations-for-femu}{%
\subsubsection{3.2.1 Essential configurations for
FEMU}\label{essential-configurations-for-femu}}

Having demonstrated the optimization capability of the DRL-DCO algorithm
on mathematical benchmark functions, the second stage of validation
applies it to FEMU for structural damage identification, using the
well-known Phase I IASC-ASCE SHM Benchmark Problem {[}48{]} as a
representative case study. The benchmark structure, shown in
\textbf{Figure 6(a)}, is a four-story, 2-bay by 2-bay steel frame with a
2.5 m by 2.5 m plan and a uniform story height of 3.5 m. This
configuration yields a 12-DOF system, with each floor possessing three
degrees of freedom, translational motion in the x and y directions and
torsional rotation about the z-axis. Among the various loading and
configuration conditions defined in the original benchmark study, this
work focuses on Case 4, in which structural asymmetry is introduced by
placing an additional 150 kg slab mass eccentrically on the top floor.
This eccentricity couples the lateral and torsional responses across all
four floors, engaging all 12 DOFs simultaneously and making the modal
identification problem considerably more demanding than the symmetric
baseline cases, providing an appropriately challenging testbed for
evaluating the robustness and accuracy of DRL-DCO in a practical FEMU
context. Further details of the benchmark structure and experimental
setup are available in Johnson et al. {[}48{]}.

Two FEMs of the 12-DOF benchmark structure are employed in this case
study, each serving a distinct and complementary role. The first is the
high-fidelity analytical FEM developed in the original benchmark study,
depicted in \textbf{Figure 6(b)}, which provides the reference modal
data against which the identification results are evaluated. This model
captures the full complexity of the structural dynamic behavior,
including coupling effects between lateral and torsional modes arising
from mass eccentricity, and features a fully defined global stiffness
matrix that faithfully represents the system's true response under both
symmetric and asymmetric conditions. A rigid diaphragm assumption is
adopted, such that each of the 4 floors is modeled with 3 DOFs, namely
lateral displacements in the x and y directions and torsional rotation
about the z-axis. The modal frequencies and mode shapes generated from
this reference model serve as the target observations for the subsequent
model updating and stiffness identification procedures.

The second model, referred to as the identification model and shown in
\textbf{Figure 6(c)}, is a simplified 12-DOF shear-type FEM that
provides a tractable parameterization for stiffness estimation. Rather
than representing the full global stiffness matrix with its coupling
terms, this model characterizes each floor \(f\) by three equivalent
stiffness parameters, \(k_{1}^{(f)}\), \(k_{2}^{(f)}\), and
\(k_{3}^{(f)}\) (\(f = 1,\ldots,4\)), representing the lateral stiffness
in the x and y directions and the torsional stiffness about the z-axis
at each inter-story level. Accordingly, a total of 12 stiffness
parameters was considered to be identified from the modal data generated
by the reference analytical FEM. This simplified modeling approach
intentionally omits inter-DOF coupling terms in the stiffness matrix in
favor of computational tractability, while preserving the essential
inter-story stiffness characteristics that are most directly indicative
of structural damage at each floor level.

% ---------- Figure 6 ----------
\begin{figure}[H]
  \centering
  \includegraphics[width=0.95\linewidth]{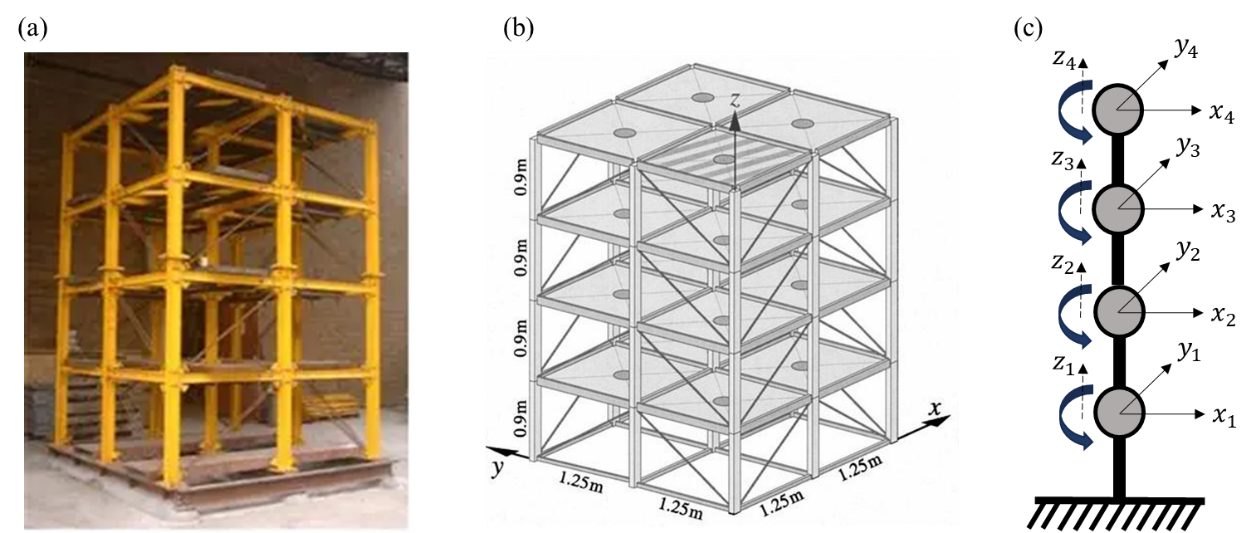}
  \caption{\textbf{(a)} Physical IASC--ASCE benchmark structure; \textbf{(b)} analytical FEM of the benchmark structure; and \textbf{(c)} simplified shear-type FEM used for stiffness parameter identification.}
\end{figure}

\hypertarget{parameter-identification-for-the-baseline-undamaged-condition}{%
\subsubsection{3.2.2 Parameter identification for the baseline
(undamaged)
condition}\label{parameter-identification-for-the-baseline-undamaged-condition}}

The first stage of this case study identifies the inter-story stiffness
parameters of the structure under its baseline (undamaged) condition,
serving as a controlled validation of DRL-DCO for FEMU prior to damage
identification. Under this condition, the true inter-story stiffness
values are uniform across all four floors, i.e., 106.6 MN/m in the lateral x-direction, 67.9 MN/m in the lateral
y-direction, and 232.02 MN/m in torsion about the z-axis, with no coupling between the three directions {[}48{]}, serving as the ground truth for this identification task.

To rigorously assess the global search capability and robustness of
DRL-DCO under highly uncertain prior knowledge, wide uniform priors were
assigned to all 12 inter-story stiffness parameters (\(k_{1}^{(f)}\),
\(k_{2}^{(f)}\), and \(k_{3}^{(f)}\) for \(f = 1,\ldots,4\)), covering
the range {[}10, 500{]} MN/m, which also served as hard bounds during
optimization. Spanning nearly two orders of magnitude around the true
values, this range intentionally simulates a practically challenging
scenario with almost no reliable prior structural information, as may
arise from degraded or undocumented structural records or significant
environmental and operational variability in real engineering settings.

Reliably extracting all 12 modal DOFs of this benchmark structure via
system identification methods such as Stochastic Subspace Identification
(SSI) is known to be practical, but also can be challenging for the
scenarios with large modeling complexity and measurement uncertainty
{[}49, 50{]}. Following the findings in {[}50{]}, only modes 1, 2, 3, 7,
9, 10, 11, and 12 can be reliably extracted via SSI, and only these
eight modes were therefore included in the FEMU objective function,
reflecting the common practical constraint of incomplete modal
information.

The FEMU objective function follows the weighted modal residual
formulation defined in Section 2.1. Convergence is declared when the
objective function value falls below a threshold of \emph{$\epsilon$} = $1\times 10^{-6}$,
which has been verified to yield sufficiently precise stiffness
identification for practical engineering purposes {[}49{]}. Further
reduction beyond this threshold is neither practically meaningful nor
physically justified given the inherent presence of measurement noise
and environmental variability in real applications.

The convergence performance of DRL-DCO in both its agent-training and
agent-inference phases is systematically summarized in \textbf{Table 6},
with representative convergence histories visualized in \textbf{Figure
7} and the corresponding identified stiffness parameters presented in
\textbf{Figure 8}. Three key observations are drawn from these results:

1) The DRL-DCO algorithm demonstrates remarkably efficient and robust
convergence on this FEMU problem. In the agent-training phase, full
convergence is achieved on average within fewer than 300 generations
with low standard deviation across independent runs. More notably,
convergence is consistently reached within the first episode of 1,000
generations, entirely within the warm-up stage, and is sustained across
all five warm-up episodes without exception. This indicates that,
although the identification problem spans a 12-dimensional parameter
space with a wide and deliberately uninformative prior range, its
optimization landscape is sufficiently well-structured for the DRL-DCO's
adaptive orchestration to reliably locate the global optimum even before
the policy has been meaningfully trained, reflecting the algorithm's
intrinsic search effectiveness independent of its RL progress.

2) In the agent-inference phase, based solely on the policy learned in a single agent-training run following the five-episode warm-up stage, DRL-DCO autonomously and consistently locates the global optimum with high precision across all independently initialized runs. More importantly, the generation-wise trajectories of the effective population size (controlled by
\(\lambda^{(t)}\)), elite ratio \(r_{\text{eli}}^{(t)}\), and orchestration ratio \(\beta^{(t)}\) exhibit notably more stable evolution during inference than during training, indicating that the learned policy has acquired a consistent search strategy rather than relying on continued online adaptation. The trained actor therefore provides reusable control decisions for different initial populations, allowing the optimization to proceed without further critic evaluation, reward feedback, or policy updates. These results demonstrate that the learned policy can retain effective search behavior after training and transfer it across independently initialized instances of the structural identification task.

3) The identified stiffness parameter values across all 12 DOFs are
consistently accurate, with mean percentage errors below 0.1\% and small
standard deviations across independent runs for all parameters. This
level of precision is achieved despite the deliberately challenging
setup of incomplete modal information involving only 8 available modes,
confirming that DRL-DCO can perform high-accuracy FEMU under practically
relevant constraints of partial observability and wide prior
uncertainty.

% ---------- Figure 7 ----------
\begin{figure}[H]
  \centering
  \includegraphics[width=0.95\linewidth]{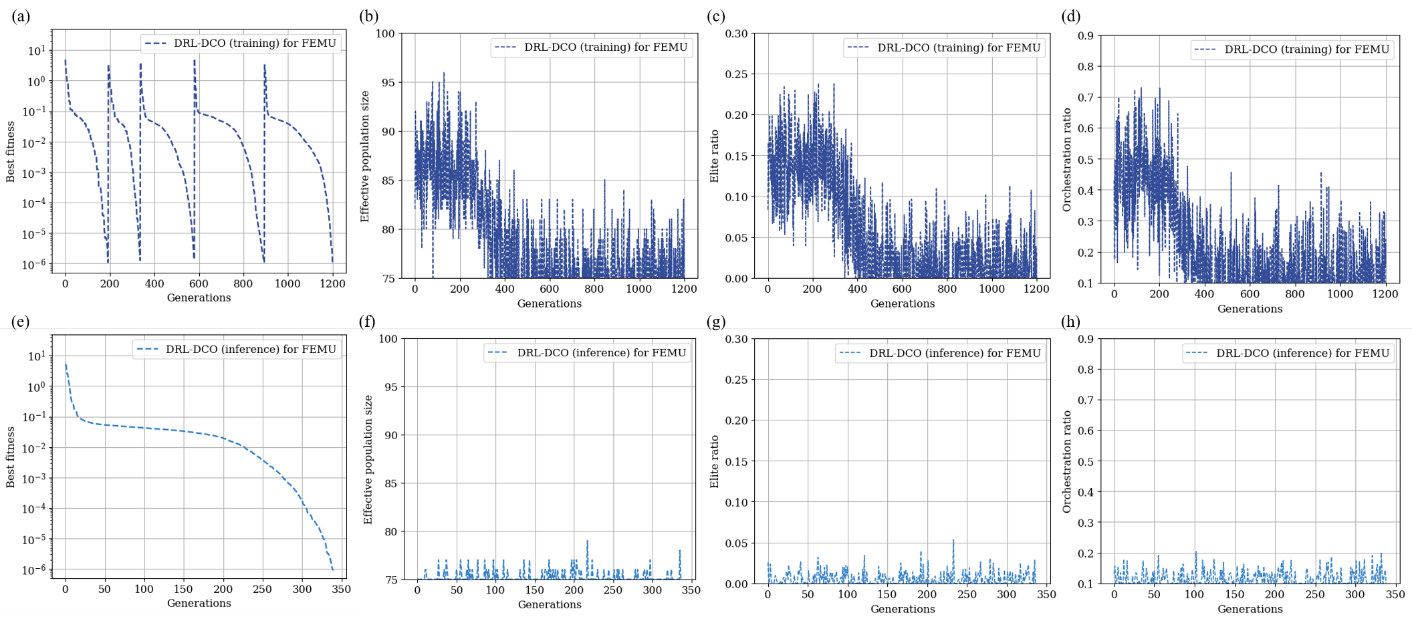}
  \caption{Convergence histories of the DRL-DCO algorithm for the FEMU on the IASC-ASCE benchmark structure, specifically regarding the best fitness, effective population size, elite ratio, and orchestration ratio, with \textbf{(a)}--\textbf{(d)} for the agent-training phase and \textbf{(e)}--\textbf{(h)} for the agent-inference phase.}
\end{figure}

% ---------- Figure 8 ----------
\begin{figure}[H]
  \centering
  \includegraphics[width=0.55\linewidth]{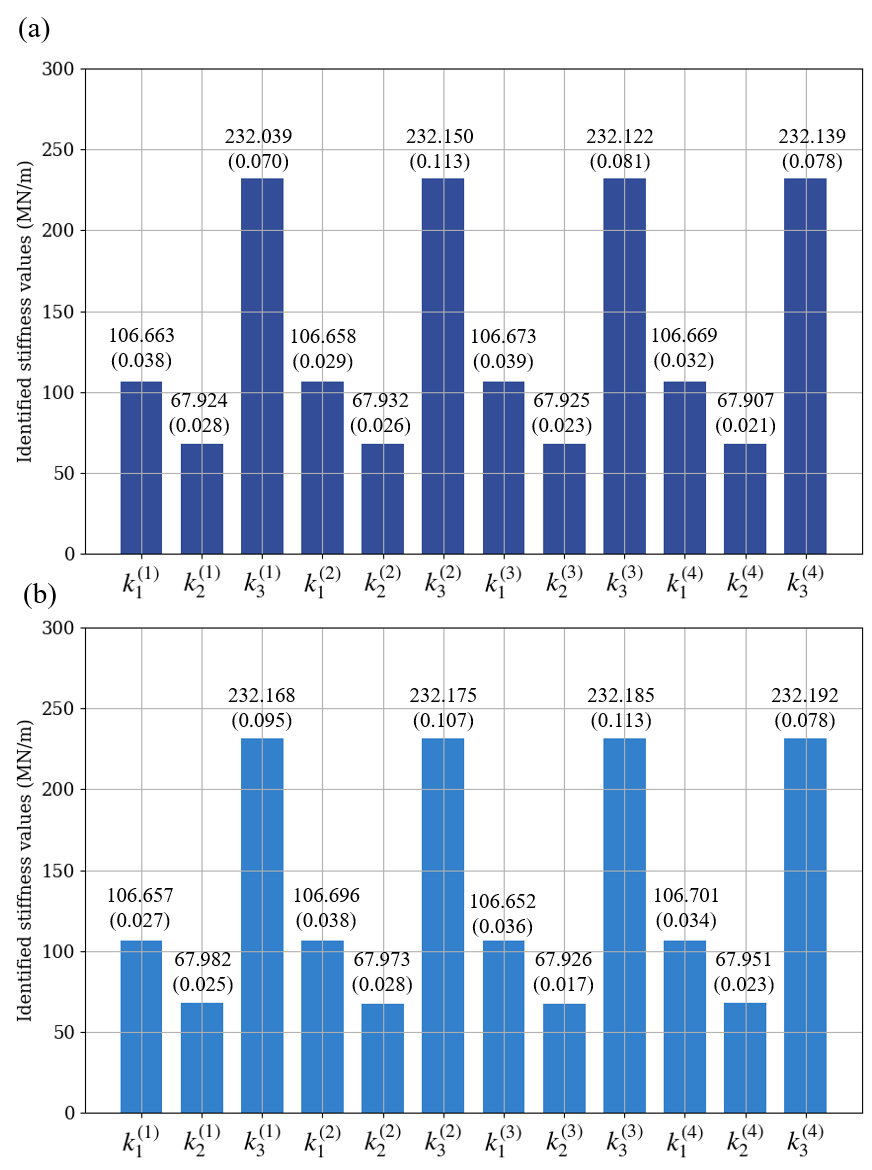}
  \caption{Identified stiffness values of the IASC-ASCE benchmark structure in its baseline condition by the DRL-DCO algorithm, with \textbf{(a)} agent-training phase results and \textbf{(b)} agent-inference phase results; values above each bar are the corresponding averages and standard deviations (in parentheses), computed across 10 independent runs.}
\end{figure}

% ---------- Table 6 ----------
\begin{table}[H]
  \caption{Convergence performance of the DRL-DCO algorithm for the FEMU of the IASC-ASCE benchmark structure's baseline condition. All results are averages over 10 independent runs, with standard deviations in parentheses.}
  \label{tab:femu}
  \centering\footnotesize\setlength{\tabcolsep}{8pt}
  \begin{tabular}{@{}lrrr@{}}
    \toprule
     & Total generations & Total evaluations & Ave.\ time (s) \\
    \midrule
    Agent-training phase  & 261 (63) & 17,642 (3,828) & 12--16 \\
    Agent-inference phase & 338 (41) & 25,176 (2,463) & 14--17 \\
    \bottomrule
  \end{tabular}
\end{table}

\hypertarget{parameter-identification-for-the-damaged-conditions}{%
\subsubsection{3.2.3 Parameter identification for the damaged
conditions}\label{parameter-identification-for-the-damaged-conditions}}

Following successful baseline identification, the DRL-DCO algorithm was
subsequently applied to identify five representative damage scenarios
defined in the original benchmark study {[}48{]} and widely adopted in
the literature {[}49{]}, summarized in \textbf{Table 7}. In all
scenarios, a broken brace is assumed to contribute zero stiffness while
the structural mass remains unchanged. The corresponding theoretical
inter-story stiffness reductions for each damage scenario, as computed
in the original study, are presented in \textbf{Table 8}.

It is noted that structural damage can in principle introduce
non-negligible coupling between lateral and torsional stiffness terms
(\(k_{1}^{(f)}\), \(k_{2}^{(f)}\), and \(k_{3}^{(f)}\) for
\(f = 1,\ldots,4\)). However, as demonstrated in the original benchmark
study {[}48{]}, the magnitudes of these off-diagonal coupling terms
remain substantially smaller than the primary diagonal stiffness
components under the considered damage scenarios. Accordingly, the
identification process retains its focus on the 12 principal stiffness
parameters, which are the most damage-sensitive and physically
interpretable quantities, while coupling terms are excluded. This
simplification helps improve computational tractability and
identification robustness, particularly under the constraint of
incomplete modal information adopted in this study.

The identified averaged results of the 12 stiffness parameters (i.e.,
\(k_{1}^{(f)}\), \(k_{2}^{(f)}\), and \(k_{3}^{(f)}\) for
\(f = 1,\ldots,4\)) across the 5 damage scenarios, achieved from the
DRL-DCO agent-inference phase, are presented in \textbf{Figure 9},
alongside the corresponding baseline-condition values and theoretical
damaged reference values for the 5 damage scenarios.

For Damage Scenarios 1 and 2, which involve complete brace removal at
the first floor and at both the first and third floors, respectively,
the DRL-DCO algorithm identifies all affected stiffness components with
close agreement to the reference values across both scenarios; for
examples, 58.43 MN/m versus 58.37 MN/m for \(k_{1}^{(1)}\), 19.81 MN/m
versus 19.67 MN/m for \(k_{2}^{(1)}\), 80.75 MN/m versus 81.30 MN/m for
\(k_{3}^{(1)}\) in Scenario 1. Additionally, the unaffected parameters
remain consistent with the baseline values. These results demonstrate
the algorithm\textquotesingle s reliability in accurately localizing and
quantifying relatively severe, multi-directional damage.

For Damage Scenarios 3 and 4, which involve single-brace failure at the
first floor and combined single-brace failures at the first and third
floors, respectively, the lateral stiffness components are identified
with reliable accuracy, while slightly larger discrepancies are observed
in the torsional components. In Scenario 3, the identified value for
\(k_{2}^{(1)}\) is 55.39 MN/m against a reference of 55.84 MN/m, while
\(k_{3}^{(1)}\) is identified as 202.21 MN/m against a reference of
209.12 MN/m. In Scenario 4, the corresponding identified values are
55.02 MN/m versus 55.84 MN/m for \(k_{2}^{(1)}\), 203.69 MN/m versus
209.12 MN/m for \(k_{3}^{(1)}\), 95.95 MN/m versus 94.54 MN/m for
\(k_{1}^{(3)}\), and 204.34 MN/m versus 210.77 MN/m for \(k_{3}^{(3)}\).
These deviations are attributable to a known limitation of the
simplified shear-type identification model: by treating lateral and
torsional stiffness as fully uncoupled, it cannot reproduce the coupled
modal interactions present in the high-fidelity benchmark FEM, whose
off-diagonal stiffness contributions are implicitly reflected in the
reference torsional reductions. This modeling discrepancy is a
structural characteristic of the identification setup rather than a
deficiency of the DRL-DCO algorithm itself, as the same discrepancy
would affect any optimizer using this identification model.

Damage Scenario 5, involving a partial one-third stiffness reduction in
a single brace at the first floor, represents the most challenging case
due to the small magnitude of the induced stiffness change. The DRL-DCO
correctly identifies the damage-affected DOFs and the direction of
stiffness reduction in all affected parameters, though with modest
numerical deviations. For examples, the identified value for
\(k_{2}^{(1)}\) is 65.71\,MN/m, slightly higher than the referenced
damaged value of 63.88\,MN/m, while the torsional component
\(k_{3}^{(1)}\) is identified as 219.79\,MN/m, lower than the referenced
value of 225.34\,MN/m. These deviations may be contributed to the
inherent sensitivity limits of reduced-order model-based identification
under low-severity damage and incomplete modal information, rather than
any convergence failure of the optimization.

Across all 5 damage scenarios, the DRL-DCO algorithm consistently
identifies the correct damage locations and produces stiffness estimates
in reasonable to close agreement with the reference values, with low
standard deviations across independent runs confirming stable and
repeatable performance. The observed discrepancies, where present, are
consistently traceable to the structural simplifications of the
identification model rather than to the optimizer's search behavior,
underscoring that the DRL-DCO algorithm performs at the practical
accuracy ceiling imposed by the identification model itself.

% ---------- Table 7 ----------
\begin{table}[H]
  \caption{Considered damage scenarios of the benchmark structure in Case Study 2.}
  \label{tab:scen}
  \centering\footnotesize\setlength{\tabcolsep}{6pt}
  \begin{tabular}{@{}ll@{}}
    \toprule
    Condition index & Damage description \\
    \midrule
    Damage Scenario 1 & All braces at the 1\textsuperscript{st} floor are broken. \\
    Damage Scenario 2 & All braces at both the 1\textsuperscript{st} and 3\textsuperscript{rd} floors are broken. \\
    Damage Scenario 3 & One brace at the 1\textsuperscript{st} floor is broken. \\
    Damage Scenario 4 & Two braces are broken, at the 1\textsuperscript{st} and the 3\textsuperscript{rd} floors. \\
    Damage Scenario 5 & One brace at the 1\textsuperscript{st} floor has a one-third reduction in stiffness. \\
    \bottomrule
  \end{tabular}
\end{table}

% ---------- Table 8 ----------
\begin{table}[H]
  \caption{Equivalent stiffness percentage loss of each DOF for the damage scenarios in Case Study 2 [48].}   % <- use your own [48] key or plain text "[48]"
  \label{tab:loss}
  \centering\footnotesize\setlength{\tabcolsep}{6pt}
  \begin{tabular}{@{}l*{5}{c}@{}}
    \toprule
    Stiffness & Scenario 1 & Scenario 2 & Scenario 3 & Scenario 4 & Scenario 5 \\
    \midrule
    $k_1^{(1)}$ & 45.24\% & 45.24\% & 0 & 0 & 0 \\
    $k_2^{(1)}$ & 71.03\% & 71.03\% & 17.76\% & 17.76\% & 5.92\% \\
    $k_3^{(1)}$ & 64.96\% & 64.96\% & 9.87\% & 9.87\% & 2.88\% \\
    $k_1^{(2)}$ & 0 & 0 & 0 & 0 & 0 \\
    $k_2^{(2)}$ & 0 & 0 & 0 & 0 & 0 \\
    $k_3^{(2)}$ & 0 & 0 & 0 & 0 & 0 \\
    $k_1^{(3)}$ & 0 & 45.24\% & 0 & 11.31\% & 0 \\
    $k_2^{(3)}$ & 0 & 71.03\% & 0 & 0 & 0 \\
    $k_3^{(3)}$ & 0 & 64.96\% & 0 & 9.16\% & 0 \\
    $k_1^{(4)}$ & 0 & 0 & 0 & 0 & 0 \\
    $k_2^{(4)}$ & 0 & 0 & 0 & 0 & 0 \\
    $k_3^{(4)}$ & 0 & 0 & 0 & 0 & 0 \\
    \bottomrule
  \end{tabular}
\end{table}

% ---------- Figure 9 ----------
\begin{figure}[H]
  \centering
  \includegraphics[width=0.85\linewidth]{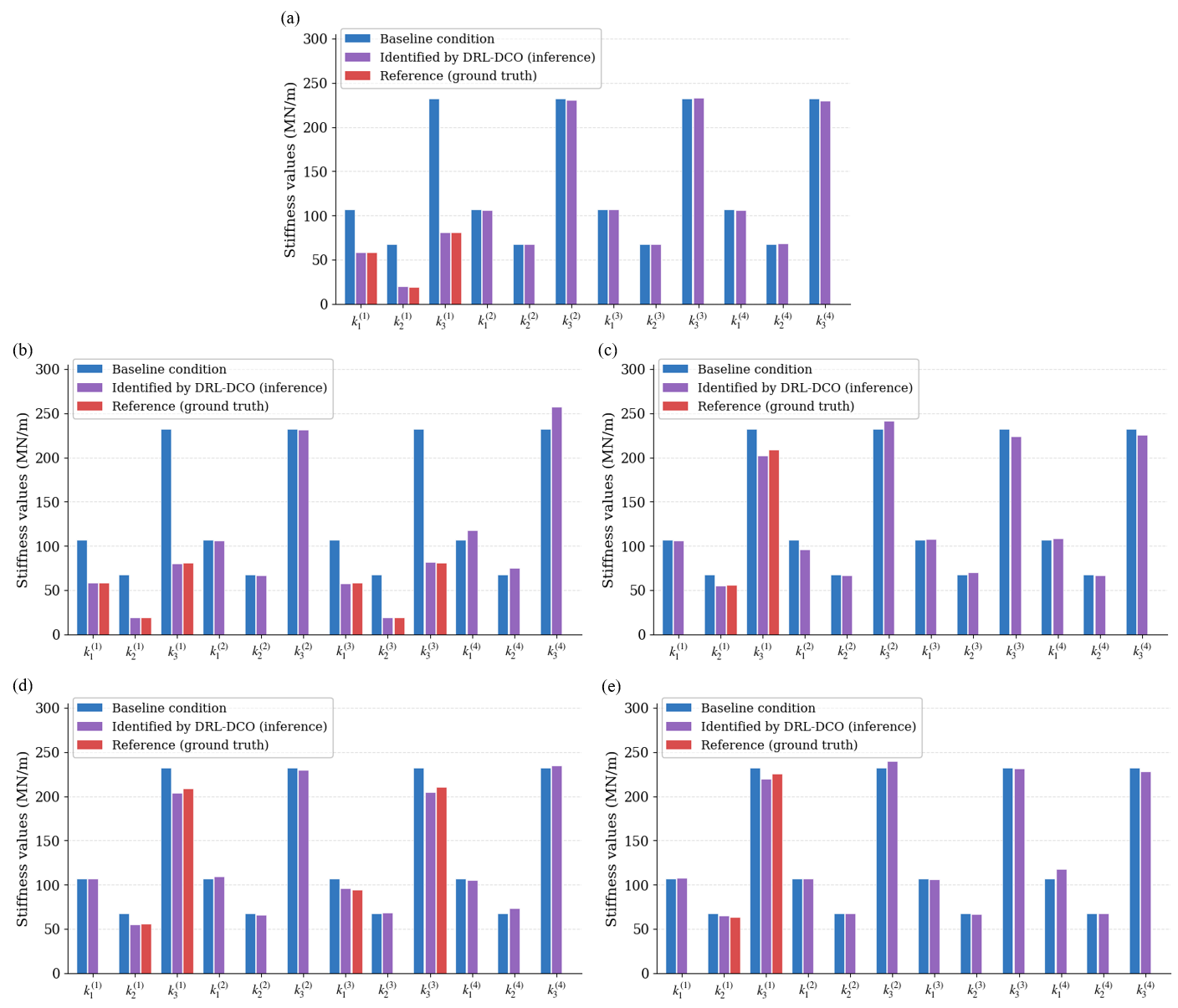}
  \caption{Identified stiffness values of the IASC-ASCE benchmark structure for the considered five damaged conditions by the DRL-DCO algorithm in its agent-inference phase, averaged across 10 independent runs, with \textbf{(a)}--\textbf{(e)} representing Damage Scenarios 1--5, respectively.}
\end{figure}

\hypertarget{conclusions}{%
\section{4. Conclusions}\label{conclusions}}

This study introduces a novel Deep Reinforcement Learning-governed dynamic DE–CMA-ES Orchestration (DRL-DCO) algorithm, built upon a Deep Deterministic Policy Gradient (DDPG)-based actor–critic agent that adaptively harmonizes Differential Evolution (DE) and Covariance Matrix Adaptation Evolution Strategy (CMA-ES) through a continuous multi-dimensional action space, a progression-aware state representation, and a diversity-informed reward function. To better address the limitations in many existing RL-assisted EA approaches, which confine the RL component to intermittent intervention on a single evolutionary strategy or leave it with limited landscape or progression awareness, and to enable supervision-free search with full automation, DRL-DCO’s actor–critic agent sustains continuous, progression-aware oversight of two evolutionary processes operating in parallel, offering a more complete realization of the RL-assisted EA paradigm. Through a critic-supervised agent-training phase and a supervision-free agent-inference phase operating via forward inference alone, the algorithm achieves robust and accurate performance across challenging high-dimensional benchmark optimization problems and demonstrates strong engineering generalization on the IASC-ASCE benchmark FEMU task. The key conclusions are as follows.

1) The proposed continuous DE--CMA-ES orchestration constitutes a
fundamental departure from conventional hard-switching hybridization.
Rather than committing the population to a single operator per
generation, DRL-DCO allocates search effort between DE and CMA-ES on the
non-elite subset of each generation through a continuously learnable
orchestration ratio, preserving the global distributional structure of
CMA-ES while retaining the diversity-restoring exploration of DE.
Ablation analysis confirms that neither DE-only nor CMA-ES-only
configurations match the full DRL-DCO across all tested problem classes,
validating the orchestration ratio as a structurally essential
component.

2) DRL-DCO features an enhanced landscape-informed state representation
that augments the candidate state vector with six progression-aware
descriptors, including fitness improvement rate, population diversity, and distributional shape statistics. This enriched
state gives the DDPG-based actor a comprehensive, temporally coherent
view of the search dynamics, enabling more informed and timely
adjustment of the DE and CMA-ES hyperparameters across generations and
episodes.

3) The incorporation of learnable population scaling and elite
preservation mechanisms enables the DRL-DCO agent to dynamically
modulate both the effective population size and the proportion of elite
solutions inherited across generations, in direct response to the
observed optimization state. This adaptive resource allocation allows
the agent to deploy smaller, computationally efficient populations
during exploitation-dominant phases and to expand search coverage during
exploration-demanding stages, improving overall efficiency and stability
on both high-dimensional nonconvex benchmark functions and the practical
FEMU task.

4) DRL-DCO incorporates a learned restart policy governed by a learnable
restart probability, estimated from stall count and diversity
statistics, which triggers population reinitialization to escape
deceptive local minima on highly multimodal landscapes such as Rastrigin
and Schwefel. This is complemented by two auxiliary, independently
learnable CMA-ES control actions, a step-size parameter and a
covariance-scaling action, which continuously regulate the search radius
and distribution orientation without discarding accumulated covariance
information.

5) DRL-DCO achieves reliable optimization performance through its two operating phases. During the agent-training phase, the actor–critic agent learns the orchestration policy through online interaction with the optimization environment and critic-mediated reward feedback. After training, the agent-inference phase uses the trained actor alone to generate generation-wise control actions through forward inference, without further policy updates or online adaptation. The stable control trajectories observed during inference indicate that the trained policy has captured effective search behavior rather than relying on continued learning during optimization. Notably, a single training run of 20–30 episodes produces a policy that generalizes reliably to independently initialized instances across all tested problem classes, demonstrating the practical reusability and self-contained intelligence of the proposed method.

Although DRL-DCO and related RL-assisted evolutionary approaches
demonstrate strong adaptability, they inevitably inherit the stochastic
uncertainty of neural network training, coupling the optimization of the
target problem with the stochastic training of the policy itself. Future
work will therefore investigate more physically informed, model-based
reinforcement learning strategies, incorporating explicit mathematical
mappings between structural parameters and target modal characteristics.
Replacing or augmenting the data-driven critic with mechanistically
transparent physics-based models and gradient-informed feedback may
further reduce training uncertainty, enhance interpretability, and
improve the reliability of RL-assisted optimization for structural
model updating and other engineering inverse problems.

\hypertarget{Acknowledgement}{%
\section{Acknowledgement}\label{Acknowledgement}}

This work was supported by the Macao Young Scholars Program (Grant No. AM2025009).

\hypertarget{references}{%
\section{References}\label{references}}

{[}1{]} Laflamme S, Blasch E, Ubertini F, Liu Z, Wertz J, Knott C, et
al. Roadmap: integrating artificial intelligence in structural health
monitoring systems. Measurement Science and Technology.
2026;37(10):103001.

{[}2{]} Li L, Brügger A, Betti R, Shen Z, Gan L, Gu H. A
cepstrum-informed neural network for vibration-based structural damage
assessment. Computers \& Structures. 2025;306:107592.

{[}3{]} Sehgal S, Kumar H. Structural dynamic model updating techniques:
a state-of-the-art review. Archives of Computational Methods in
Engineering. 2016;23(3):515-33.

{[}4{]} Ereiz S, Duvnjak I, Jiménez-Alonso JF. Review of finite element
model updating methods for structural applications. Structures.
2022;41:684-723.

{[}5{]} Mottershead JE, Link M, Friswell MI. The sensitivity method in
finite element model updating: a tutorial. Mechanical Systems and Signal
Processing. 2011;25(7):2275-96.

{[}6{]} Alkayem NF, Cao M, Zhang Y, Bayat M, Su Z. Structural damage
detection using finite element model updating with evolutionary
algorithms: a survey. Neural Computing and Applications.
2018;30(2):389-411.

{[}7{]} Holland JH. Adaptation in natural and artificial systems: an
introductory analysis with applications to biology, control, and
artificial intelligence. Cambridge, MA: MIT Press; 1992.

{[}8{]} Kennedy J, Eberhart R. Particle swarm optimization. In:
Proceedings of the 1995 IEEE International Conference on Neural
Networks. Piscataway, NJ: IEEE; 1995. p. 1942-8.

{[}9{]} Price KV. Differential evolution. In: Handbook of Optimization:
From Classical to Modern Approach. Berlin, Heidelberg: Springer; 2013.
p. 187-214.

{[}10{]} Auger A, Hansen N. Tutorial CMA-ES: evolution strategies and
covariance matrix adaptation. In: Proceedings of the 14th Annual
Conference Companion on Genetic and Evolutionary Computation. New York:
ACM; 2012. p. 827-48.

{[}11{]} Li L, Shen Z, Gan L, Xu L, Zhang H, Ye W. A novel
cepstrum-based and evolutionary optimization hybrid framework for
structural parameter identification. Advances in Structural Engineering.
2025;13694332251405899.

{[}12{]} Črepinšek M, Liu SH, Mernik M. Exploration and exploitation in
evolutionary algorithms: a survey. ACM Computing Surveys.
2013;45(3):1-33.

{[}13{]} Lye A, Cicirello A, Patelli E. Sampling methods for solving
Bayesian model updating problems: a tutorial. Mechanical Systems and
Signal Processing. 2021;159:107760.

{[}14{]} Beck JL, Au SK. Bayesian updating of structural models and
reliability using Markov chain Monte Carlo simulation. Journal of
Engineering Mechanics. 2002;128(4):380-91.

{[}15{]} Ching J, Chen YC. Transitional Markov chain Monte Carlo method
for Bayesian model updating, model class selection, and model averaging.
Journal of Engineering Mechanics. 2007;133(7):816-32.

{[}16{]} Chatzi EN, Smyth AW. The unscented Kalman filter and particle
filter methods for nonlinear structural system identification with
non-collocated heterogeneous sensing. Structural Control and Health
Monitoring. 2009;16(1):99-123.

{[}17{]} Li Q, Ni P, Du X, Han Q, Xu K, Bai Y. Bayesian finite element
model updating with a variational autoencoder and polynomial chaos
expansion. Engineering Structures. 2024;316:118606.

{[}18{]} Itoi T, Amishiki K, Lee S, Yaoyama T. Bayesian structural model
updating with multimodal variational autoencoder. arXiv:2406.09051
{[}preprint{]}. 2024.

{[}19{]} Samadian D, Muhit IB, Dawood N. Application of data-driven
surrogate models in structural engineering: a literature review.
Archives of Computational Methods in Engineering. 2025;32(2):735-84.

{[}20{]} Kudela J, Matousek R. Recent advances and applications of
surrogate models for finite element method computations: a review. Soft
Computing. 2022;26(24):13709-33.

{[}21{]} Sabamehr A, Lim C, Bagchi A. System identification and model
updating of highway bridges using ambient vibration tests. Journal of
Civil Structural Health Monitoring. 2018;8:755-71.

{[}22{]} Yin T, Zhu HP. An efficient algorithm for architecture design
of Bayesian neural network in structural model updating. Computer-Aided
Civil and Infrastructure Engineering. 2020;35(4):354-72.

{[}23{]} Zhang Z, Sun C. Structural damage identification via
physics-guided machine learning: a methodology integrating pattern
recognition with finite element model updating. Structural Health
Monitoring. 2021;20(4):1675-88.

{[}24{]} Cuomo S, Di Cola VS, Giampaolo F, Rozza G, Raissi M, Piccialli
F. Scientific machine learning through physics-informed neural networks:
where we are and what\textquotesingle s next. Journal of Scientific
Computing. 2022;92(3):88.

{[}25{]} Zhou K, Tang J. Structural model updating using adaptive
multi-response Gaussian process meta-modeling. Mechanical Systems and
Signal Processing. 2021;147:107121.

{[}26{]} Hong F, Wei P, Bi S, Beer M. Efficient variational Bayesian
model updating by Bayesian active learning. Mechanical Systems and
Signal Processing. 2025;224:112113.

{[}27{]} Bruno G, Parisi F, Ruggieri S, Uva G. An intelligent
agent-based method for model updating of structures with multiple
uncertain parameters via deep reinforcement learning. Mechanical Systems
and Signal Processing. 2025;234:112832.

{[}28{]} Li SE. Deep reinforcement learning. In: Reinforcement Learning
for Sequential Decision and Optimal Control. Singapore: Springer Nature;
2023. p. 365-402.

{[}29{]} Meng X, Zhang H, Jia K, Li H, Huang Y. Optimal sensor placement
for structural health monitoring based on deep reinforcement learning.
Smart Structures and Systems. 2023;31(3):247-57.

{[}30{]} Zhong Q, Wei S, Huang Y, Jia K, Li H. Optimal placement of
multiple types of sensors based on Bayesian system identification using
heuristic-enhanced deep reinforcement learning. Engineering Structures.
2025;338:120570.

{[}31{]} Guo X, Lin P, Wang Q, Hu G. Intelligent control of structural
vibrations based on deep reinforcement learning. Journal of
Infrastructure Intelligence and Resilience. 2025;4(2):100136.

{[}32{]} Nishat TA, Jeong JH, Jo H, Xia S, Liu J. Reinforcement learning
for adaptive battery management of structural health monitoring IoT
sensor network. Applied Energy. 2025;390:125731.

{[}33{]} Bai H, Cheng R, Jin Y. Evolutionary reinforcement learning: a
survey. Intelligent Computing. 2023;2:0025.

{[}34{]} Li P, Hao J, Tang H, Fu X, Zhen Y, Tang K. Bridging
evolutionary algorithms and reinforcement learning: a comprehensive
survey on hybrid algorithms. IEEE Transactions on Evolutionary
Computation. 2024.

{[}35{]} Gangwani T, Peng J. Policy optimization by genetic
distillation. arXiv:1711.01012 {[}preprint{]}. 2017.

{[}36{]} Sumiea EH, Abdulkadir SJ, Alhussian HS, Al-Selwi SM, Alqushaibi
A, Ragab MG, et al. Deep deterministic policy gradient algorithm: a
systematic review. Heliyon. 2024;10(9).

{[}37{]} Filar J, Vrieze K. Competitive Markov decision processes. New
York: Springer; 2012.

{[}38{]} Zhang J, Sanderson AC. JADE: adaptive differential evolution
with optional external archive. IEEE Transactions on Evolutionary
Computation. 2009;13(5):945-58.

{[}39{]} Tanabe R, Fukunaga AS. Improving the search performance of
SHADE using linear population size reduction. In: 2014 IEEE Congress on
Evolutionary Computation (CEC). Piscataway, NJ: IEEE; 2014. p. 1658-65.

{[}40{]} Hansen N. The CMA evolution strategy: a comparing review. In:
Towards a New Evolutionary Computation: Advances in the Estimation of
Distribution Algorithms. Berlin, Heidelberg: Springer; 2006. p. 75-102.

{[}41{]} Jamil M, Yang XS. A literature survey of benchmark functions
for global optimisation problems. International Journal of Mathematical
Modelling and Numerical Optimisation. 2013;4(2):150-94.

{[}42{]} Wolpert DH, Macready WG. No free lunch theorems for
optimization. IEEE Transactions on Evolutionary Computation.
1997;1(1):67-82.

{[}43{]} Suganthan PN, Hansen N, Liang JJ, Deb K, Chen YP, Auger A,
Tiwari S. Problem definitions and evaluation criteria for the CEC 2005
special session on real-parameter optimization. KanGAL Report 2005005.
2005.

{[}44{]} Mohamed AW, Hadi AA, Fattouh AM, Jambi KM. LSHADE with
semi-parameter adaptation hybrid with CMA-ES for solving CEC 2017
benchmark problems. In: 2017 IEEE Congress on Evolutionary Computation
(CEC). Piscataway, NJ: IEEE; 2017. p. 145-52.

{[}45{]} Hansen N, Kern S. Evaluating the CMA evolution strategy on
multimodal test functions. In: Parallel Problem Solving from Nature -
PPSN VIII. Berlin, Heidelberg: Springer; 2004. p. 282-91.

{[}46{]} Auger A, Hansen N. A restart CMA evolution strategy with
increasing population size. In: 2005 IEEE Congress on Evolutionary
Computation. Piscataway, NJ: IEEE; 2005. p. 1769-76.

{[}47{]} Hansen N, Ros R, Mauny N, Schoenauer M, Auger A. Impacts of
invariance in search: when CMA-ES and PSO face ill-conditioned and
non-separable problems. Applied Soft Computing. 2011;11(8):5755-69.

{[}48{]} Johnson EA, Lam HF, Katafygiotis LS, Beck JL. Phase I IASC-ASCE
structural health monitoring benchmark problem using simulated data.
Journal of Engineering Mechanics. 2004;130(1):3-15

{[}49{]} Das S, Saha P. Structural health monitoring techniques
implemented on IASC-ASCE benchmark problem: a review. Journal of Civil
Structural Health Monitoring. 2018;8(4):689-718.

{[}50{]} Cara FJ, Carpio J, Juan J, Alarcón E. An approach to
operational modal analysis using the expectation maximization algorithm.
Mechanical Systems and Signal Processing. 2012;31:109-29.

\end{document}